%% file: main.tex
\documentclass{article}

\input{macro.tex}

\usepackage{microtype}
\usepackage{graphicx}
\usepackage{subfigure}
\usepackage{booktabs} 

\usepackage{hyperref}

\usepackage[accepted]{icml2023}

\usepackage{amsmath}
\usepackage{amssymb}
\usepackage{mathtools}
\usepackage{amsthm}

\usepackage[capitalize,noabbrev]{cleveref}

\theoremstyle{plain}
\theoremstyle{definition}

\theoremstyle{remark}

\icmltitlerunning{Performative Recommendation: Diversifying Content via Strategic Incentives}

\begin{document}

\twocolumn[
\icmltitle{Performative Recommendation: Diversifying Content via Strategic Incentives}



\icmlsetsymbol{equal}{*}

\begin{icmlauthorlist}
\icmlauthor{Itay Eilat}{technion}
\icmlauthor{Nir Rosenfeld}{technion}
\end{icmlauthorlist}

\icmlaffiliation{technion}{Faculty of Computer Science, Technion -- Israel Institute of Technology, Haifa, Israel}

\icmlcorrespondingauthor{Nir Rosenfeld}{nirr@cs.technion.ac.il}

\icmlkeywords{Machine Learning, ICML}

\vskip 0.3in
]



\printAffiliationsAndNotice{}  

\begin{abstract}
\input{abstract}
\end{abstract}

\todo{for camera ready, change to `accepted' to add ICML notice string}

\input{intro}

\input{related}

\input{learning_setup}

\input{method}

\input{analysis}

\input{synthetic}
\input{real_data_exp}

\input{Discussion}

\subsection*{Acknowledgements}
This research was supported by the Israel Science Foundation (grant No. 278/22) and by VATAT Fund to the Technion Artificial Intelligence Hub (Tech.AI). 

\bibliography{refs.bib}
\bibliographystyle{icml2023}

\newpage
\appendix
\onecolumn
\input{appendix}


\end{document}

%% file: macro.tex
\usepackage{verbatim}
\usepackage{microtype}
\usepackage{subfigure}
\usepackage{enumitem}
\usepackage{booktabs} 
\usepackage{color,graphicx}
\usepackage{comment}
\usepackage{thmtools, thm-restate}
\usepackage{hyperref,xcolor}
\usepackage{hyperref}
\usepackage{amssymb,amsmath,amsthm,dsfont,bm,mathtools, bbm}
\usepackage{xfrac}
\usepackage[bottom]{footmisc}

\newcommand{\red}[1]{} 
\newcommand{\blue}[1]{} 
\newcommand{\green}[1]{} 
\newcommand{\purple}[1]{} 
\newcommand{\orange}[1]{} 

\newcommand\todo[1]{{\red{TODO: {#1}}}}

\newcommand\tocitec[1]{\red{[CITE: {#1}]}}
\newcommand\toref{\red{[REF]}}

\newcommand\extended[1]{{\orange{EXTENDED: {#1}}}}

\DeclareMathOperator*{\argmax}{argmax}

\DeclareMathOperator{\rev}{rev}

\DeclareMathOperator{\rank}{rank}

\DeclareMathOperator{\avg}{avg}

\DeclareMathOperator{\diversity}{div}
\newcommand{\divat}[1]{{\mathrm{div}_{@{#1}}}}
\newcommand{\divk}{\divat{k}}

\newcommand{\DCGk}{\ensuremath{\mathrm{DCG}_{@k}}}
\newcommand{\NDCGat}[1]{\ensuremath{\mathrm{NDCG}_{@{#1}}}}
\newcommand{\NDCGk}{\NDCGat{k}}

\newcommand{\method}[1]{{\fontfamily{lmtt}\selectfont{{#1}}}} 

\newcommand{\1}[1]{\mathds{1}{\{{#1}\}}}

\renewcommand{\r}{{\bm{r}}}
\newcommand{\rhat}{{\hat{\r}}}
\newcommand{\y}{{\bm{y}}}

\newcommand{\R}{{\mathbb{R}}}

\newcommand{\vtilde}{{\tilde{v}}}

\newcommand{\Phat}{{\hat{P}}}

\newtheorem{theorem}{Theorem}

\newtheorem{lemma}[theorem]{Lemma}

\newtheorem{proposition}[theorem]{Proposition}

%% file: abstract.tex
The primary goal in recommendation is to suggest relevant content to users,
but optimizing for accuracy often results in recommendations that lack diversity.
To remedy this, conventional approaches
such as re-ranking improve diversity
by \emph{presenting} more diverse items.
Here we argue that to promote inherent and prolonged diversity, the system must encourage its \emph{creation}.
Towards this, we harness the performative nature of recommendation,
and show how learning can incentivize
strategic content creators to create diverse content.
Our approach relies on a novel form of regularization that anticipates strategic changes to content, and penalizes for content homogeneity.
We provide analytic and empirical results that demonstrate when and how diversity can be incentivized,
and experimentally demonstrate the utility of our approach on synthetic and semi-synthetic data.


%% file: intro.tex
\section{Introduction}
\label{sec:intro}


Recommendation has become a key driving force in determining what content we are exposed to,
and ultimately, which we consume \citep{mackenzie2013retailers,ursu2018power}.
But despite the commercial success of modern recommendation systems,
a known shortcoming is that recommendations tend to be insufficiently diverse,
with content homogeneity becoming more pronounced over time;
this has been a longstanding issue in the field for over two decades \citep{carbonell1998use,bradley2001improving}.
Diversity is important in recommendation not only for improving recommendation quality \citep{vargas2014improving,kaminskas2016diversity}
and user satisfaction \citep{herlocker2004evaluating,ziegler2005improving,mcnee2006being,hu2011helping,wu2018personalizing,dean2020recommendations},
but also because a lack of diversity can lead to inequity across content creators,
which often hurts the `long-tail' of non-mainstream suppliers
\citep{yin2012challenging,burke2017multisided,singh2018fairness,abdollahpouri2019unfairness,mladenov2020optimizing,wang2021user}.
From the perspective of the recommendation platform,
an inability to diversify content translates into an inability
to utilize the full potential that lies in
the natural variation of user preferences
for promoting system goals
\citep{anderson2006long,yin2012challenging}.
This has lead to widespread interest in developing methods for  making recommendations more diverse \citep{kunaver2017diversity}. 

The common approach for diversifying recommendations
is to apply some post-processing procedure 
that re-ranks the output of a conventionally-trained ranking model---which is optimized for predicting user-item relevance---to be more diverse,
for example by traversing the ranked list and removing items which are similar to higher-ranked items
\citep{carbonell1998use,ziegler2005improving,sha2016framework}.
This simple heuristic approach has been shown to be quite effective---at least when considering a \emph{given} ranked list, and at one point in time.
But recommendation is inherently a dynamic process:
here we argue that post-hoc methods may not suffice for promoting diversity in the long run.\looseness=-1

To see why, 
consider that re-ranking (and similar approaches)
are designed to diversify the \emph{presentation} of content---not content itself.
Presenting diverse content may help in the specific instance it targets,
but does not change the pool of available items,
nor does it account for any downstream affects of recommendation.
Recent work has shown that one drawback of using prediction as the basis for recommendation is that it causes homogenization \citep{chaney2018algorithmic}\todo{add more};
here we argue that rearranging predicted items to form the appearance of diversity does not remedy this.\looseness=-1

As an alternative, we propose to encourage the \emph{creation} of diverse content, so that 
the set of available items becomes inherently more diverse.
In this way, we aim to target the cause---rather than the symptom.
Our main observation is that content is shaped by \emph{content creators}, who seek to maximize exposure to their items \citep{ben2020content,hron2022modeling},
and hence likely to modify content in ways which promote their item's predicted relevance (or `score').
Since the goal of learning is to infer such scores,
this gives the learning system leverage in shaping the 
incentives of content creators.
Here we propose to utilize this power to incentivize
for the creation of more diverse content.

Towards this, 
we draw connections to the related literature on strategic 
learning 
\citep{bruckner2012static,hardt2016strategic},
and model content creators as gaining utility from the score given to their items by the learned predictive model.
Content creators can then improve their utility
by strategically modifying their items to obtain a higher score. 
Thus, a learned predictive model determines not only what items are recommended to which users---but also creates incentives for content creators,
which can promote change
\citep{ben2018game,jagadeesan2022supply}.
This provides the system with the potential power to steer
the collection of renewing items---over time, 
and with proper incentivization---towards diversity \citep{hardt2022performative}.


To study when and how the system can effectively exercise its power,
we cast recommendation as an instance of
\emph{performative prediction} \citep{perdomo2020performative},
which subsumes and extends strategic learning to a temporal setting where
repeated learning causes the underlying data to shift over time.
Focusing on retraining dynamics,
we study when and how learning
can be used to incentivize the creation and preservation of diversity.
In retraining, our only means for driving incentivizes derives from 
\emph{how} we retrain, i.e., from our criterion for choosing the predictive model at each round.
Since retraining aims for models that are predictively accurate,
our goal will be to provide recommendations that are accurate \emph{and} diverse.
But diversity and accuracy can be at odds;
hence, we seek to understand how they relate, 
and to propose ways in which their tradeoff can be optimally exploited.\looseness=-1



We begin with a basic analysis demonstrating the mechanisms through which incentives and diversity relate within our setup.
We then propose a learning objective that allows to balance ranking accuracy
(and in particular NDCG) with diversity through a novel form of regularization,
which we use to maximize diversity under accuracy constraints.
Our proposed diversity regularizer has two main benefits.
First, it is differentiable,
and hence can be optimized using gradient methods.
Second, it can be applied to \emph{strategically-modified inputs};
this equips our objective with the ability to anticipate the
strategic responses of content creators,
and hence, to encourage predictive rules that incentivize diversity.
Our proposed strategic response operator is also differentiable;
thus, and using recent advances in differentiable learning-to-rank,
our entire strategic learning objective becomes differentiable,
and can be efficiently optimized end-to-end.

Using our proposed learning framework, we empirically demonstrate
how properly accounting for strategic incentives can improve diversity---and how neglecting to do so can lead to homogenization.
We begin with a series of synthetic experiments, each designed to study a different aspect of our setup, such as the role of time,
the natural variation in user preferences,
and the cost of applying strategic updates.
We then evaluate our approach in a semi-synthetic environment 
using real data (Yelp restaurants) and simulated responses.
Our results demonstrate the ability of strategically-aware retraining to bolster diversity,
and illustrate the importance of incentivizing the creation of diversity.
All code is made publicly available at:
\url{https://github.com/itayeilat/Performative-Recommendation}.

\todo{for camera change to public repo}

\todo{add illustration of dynamics (=updates+retraining) for myopic vs. strategic}

\red{
\paragraph{Contributions.}
Our work makes the following contributions:
\begin{itemize}
\item ...
\end{itemize}
}


%% file: related.tex
\subsection{Related work}
\label{sec:related}


\textbf{Diversity in recommendation.}\,\,
The literature on diversity in recommendation is extensive; here we present a relevant subset.
Early approaches propose to diversify via re-ranking
\citep{carbonell1998use, bradley2001improving, ziegler2005improving},
an approach that remains to be in widespread use today
\citep{abdollahpouri2019managing}.
More recent methods include 
diversifying via functional optimization \citep{zhang2008avoiding}
or integration within matrix factorization \citep{su2013set, hurley2013personalised,cheng2017learning}.
Diversity has also been studied in
sequential \citep{kim2019sequential},
conversational \citep{fu2021popcorn},
and adversarial bandit \citep{browndiversified}
settings.
%
The idea of using regularization to promote secondary objectives
in recommendation has been applied for
controlling popularity bias \citep{abdollahpouri2017controlling},
enhancing neutrality \citep{kamishima2014correcting},
and promoting equal opportunity \citep{zhu2021popularity}.
For diversity,
\citet{wasilewski2016incorporating} apply regularization,
but assume that the system has direct control over (latent) item features;
this is crucially distinct from our setting in which
the system can only indirectly encourage content creators to apply changes.


\textbf{Strategic learning.}\,\,
There has been much recent interest in studying learning in the presence of
strategic behavior.
\citet{hardt2016strategic} propose \emph{strategic classification} as a framework
for studying classification tasks in which users (who are the targets of prediction) can modify their features---at a cost---to obtain favorable predictions.
This is based on earlier formulations by
\citet{bruckner2009nash,bruckner2012static},
with recent works extending the framework to settings in which users act on noisy \citep{jagadeesan2021alternative}
or missing information \cite{ghalme2021strategic,bechavod2022information},
have broader interests \citep{levanon2022generalized},
or are connected by a graph \citep{eilat2022strategic}.
Since we model content creators as responding to a scoring rule,
our work pertains to the subliterature on \emph{strategic regression},
in which user utility derives from a continuous function
\citep{rosenfeld2020predictions,tang2021linear,harris2021stateful,bechavod2022information},
and strategic behavior is often assumed to also affect outcomes
\citep{shavit2020causal,harris2022strategic}.
Within this field, our framework is unique in that it considers content creators---rather than end-users---as the focal strategic entities.
The main distinction is that this requires learning to account for the \emph{joint} behavior of all strategic agents (rather than each individually),
which even for linear score functions results in complex behavioral patters 
(c.f. standard settings in which linearity implies uniform movement \citep{liu2022strategic}).
Regularization has been used to control incentives
in \citet{rosenfeld2020predictions,levanon2021strategic},
but in distinct settings and towards different goals (i.e., unrelated to recommendation or diversity),
and for user responses that fully decompose.\looseness=-1

\textbf{Performativity and incentives.}\,\,
The current literature on performative learning focuses primarily on macro-level analysis,
such as providing sufficient global conditions for retraining to converge
\citep{perdomo2020performative,miller2021outside,brown2022performative}
or proposing general optimization algorithms
\citep{mendler2020stochastic,izzo2021learn,drusvyatskiy2022stochastic,maheshwari2022zeroth}.
In contrast, performativity in our setting emerges from micro-level modeling of strategic agents in a dynamic recommendation environment,
and our goal is to address the specific challenges inherent in our focal learning task.
Within recommendation,
content creators (or `supplier') incentives 
have also been studied from a game-theoretic perspective 
\citep{ben2018game,Ben-Porat_Goren_Rosenberg_Tennenholtz_2019,ben2020content,jagadeesan2022supply,hron2022modeling}.
Here, focus tends to be on notions of equilibrium,
and the system is typically assumed to have direct control over outcomes
(e.g., determining allocations or monetary rewards).
Our work focuses primarily on learning,
and studies indirect incentivization through a learned predictive rule.

\extended{
Finally, we note that in terms of experimentation,
virtually all works that study behavior in recommendation---as does ours---rely on simulation for evaluation
\citep[e.g.,][]{jiang2019degenerate,sun2019debiasing,zhan2021towards,kalimeris2021preference,yao2021measuring},
and in particular for retraining \citep{krauth2020offline}.
Recommendation is at its core a policy problem;
hence, appropriate evaluation requires means for measuring counterfactual outcomes
(i.e., what ``would have happened'' under a different recommendation)---for which simulation has been advocated as a primary candidate
\citep{ie2019recsim,chaney2021recommendation,curmei2022towards}.
}

%% file: learning_setup.tex
\section{Problem Setup}
\label{sec:Setup}

Our setup considers a recommendation platform consisting of $m$ users and $n$ items.
Items are described by feature vectors $x_j \in \R^d$, $j \in [n]$,
and each item $x_j$ is owned by a (strategic) content creator $j$.
The goal of the system is to learn latent vector representations $u_i \in \R^d$
for each user $i \in [m]$ that are useful for recommending relevant items.
As in \citet{hron2022modeling}, we assume all features
are constrained to have unit $\ell_2$ norm, $\|u\| = \|x\| = 1$
(i.e., lie on the unit sphere).
This ensures equal treatment across items (by the system)
and users (by content creators),
and prevents features from growing indefinitely due to strategic updates.

The system makes recommendations by ranking items for each user $i$ using a personalized score function $f_i(x)$ as:
\begin{equation}
\r_i = \rank(f_i(x_1),\dots,f_i(x_n))
\,\,\quad \forall i \in [m]
\end{equation}
where $f_i(x) = f(x;u_i)$ rely on learned user representation vectors $u_i$.
For a list of items $X$ we denote in shorthand $\r_i=\rank(f(X;u_i))$.
As in most works on strategic learning
\citep[e.g.,][]{hron2022modeling, jagadeesan2022supply, carroll2022estimating},
we consider linear score functions $f(x;u_i) = u_i^\top x$.
Overall, the goal of the system is to learn good $f_1,\dots,f_n$ from data,
where $u_1,\dots,u_n$ are the learned parameters.

\extended{Linear scores are used extensively in the strategic learning literature
since the learned $u$ can be interpreted as estimated user preferences
(in line with expected utility theory \citep{mas1995microeconomic}),
and since they permit tractable strategic responses. 
In recommendation, linear models have been shown to be sufficiently expressive in multiple settings \citep{ferrari2019we}.
Despite the linearity of $f$,
our learning problem is highly non-linear 
due to norm constraints, strategic responses,
and the ranking objective.}

\textbf{Learning objective.}\,\,
We measure ranking quality using the standard measure of NDCG evaluated on the top $k$ items,
defined as follows.
Consider a list of items with relevance scores $\y=(y_1,\dots, y_n)$.
Let $\r=(r_1,\dots,r_n)$ be a ranking,
and denote by $\r(\ell)$ the index of the $\ell^{\text{th}}$-ranked item in $\r$.
Then the top-$k$ discounted cumulative gain (DCG) is: 
\begin{equation}
\label{eq:DCG@k}
\DCGk(\y, \r) = \sum\nolimits_{\ell=1}^k\frac{2^{y_{\r(\ell)}}-1}{\log(1+\ell)}
\end{equation}
where $2^{y_j}-1$ measures the `gain' in relevance from having item $j=r(\ell)$ in the top $k$, and $\log(1+\ell)$ `discounts' its rank.
NDCG is then obtained by normalizing relative to the optimal ranking
$\r^* = \argmax_{\r} \DCGk(\y,\r)$.
The primary goal of the system is therefore to learn
user representations $\{u_i\}_{i=1}^n$
that optimize average top-$k$ NDCG:
\begin{equation}
\max_{u_1,\dots,u_m} \frac{1}{m} \sum\nolimits_{i=1}^m
\NDCGk(\y_i,\r_i)
\end{equation}
where $\r_i$ is the ranking of items for user $i$ according to $f(x;u_i)$.
For learning, we will assume that the system has access to relevance labels
$y_{ij}$ for some user-item pairs $(i,j)$,
and the goal is to generalize well to other pairs.

\textbf{Item diversity.}\,\,
In addition to ranking accuracy,
we will also be interested in measuring and promoting diversity across recommended items.
Here we consider intra-list diversity
\citep{ziegler2005improving,vargas2011rank,antikacioglu2019new}
and focus primarily on cosine similarity as a metric,
$\cos(x,x') = \frac{x^\top x'}{\|x\|\|x'\|}$, 
which is appropriate for comparing unit-norm features
\citep{ekstrand2014user, hron2022modeling}.%
\footnote{Since features are normalized, we have $\cos(x,x') = x^\top x'$.}
See Appendix~\ref{apx:entropy} for an extension of our approach to entropy-based similarity.

For a list of items $X=(x_1,\dots,x_K)$ and corresponding ranking $\r$,
diversity for the top-$k$ items is defined as:
\begin{equation}
\label{eq:divk}
\divk(X,\r) = 
\frac{1}{k(k-1)}
\sum\nolimits_{j,\ell=1}^k 1-\cos(x_{\r(j)},x_{\r(\ell)})
\end{equation}
which takes values in $[0,1]$ \citep{smyth2001similarity}.

\textbf{Recommendation graph.}\,\,
In our setting, each user $i$ is associated with a list of $K_i$ potentially-relevant items, 
denoted $X_i \subseteq [n]$, 
and the system's goal is to choose a subset of $k\le K_i$ items to recommend as a ranked list.\footnote{This is also known as 
second-stage recommendation; see e.g. \citet{ma2020off,hron2021component,wang2022fairness}.}
Since the same items can appear in multiple lists,
it will be useful to consider users and items through a bipartite graph $G=(U,X,E)$, where $(i,j) \in E$ if item $j$ is in user $i$'s list of candidate items.
\extended{\footnote{For simplicity we assume $G$ is fixed, but note our framework also supports time-varying graphs.}}
We will also denote $U_j = \{ i \in [m] \mid (i,j) \in E \}$
and $m_j = |U_j|$.
As we will see, the graph plays a key role in determining the system's potential for encouraging diversity.


\begin{figure*}[t]
	\centering{
	\includegraphics[height=3.8cm]{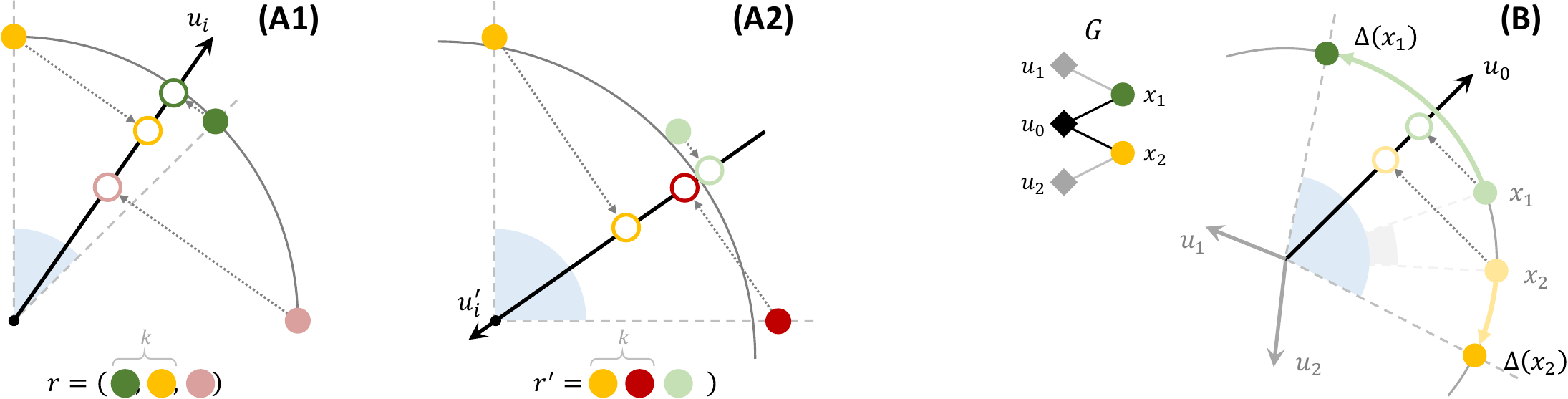}
	}
	\caption{
	\textbf{(A)} \textbf{Tradeoff between accuracy and diversity:}
	Consider three items with high ({\color[RGB]{84,130,53}$\huge{\bullet}$}),  
	medium ({\color[RGB]{255,192,0}$\huge{\bullet}$}), and low
	({\color[RGB]{192,0,0}$\huge{\bullet}$}) relevance.
	Learned user embedding $u_i$ obtains high $\NDCGat{2}$ (\textit{projected points}) but low $\divat{2}$ (\textit{blue sector}) 
	\textbf{(A1)}.
	Conversely, $u'_i$ (note flipped orientation)
	obtains high diversity, but at the cost of reduced accuracy \textbf{(A2)}.
	\textbf{(B)} \textbf{Creating diversity with strategic incentives:}
	Consider two items 
	and three users 
	connected by the graph $G$.
	Initially, both items are similar (\textit{light circles}), and diversity is low. However, $u_1$ and $u_2$ incentivize $x_1$ and $x_2$, respectfully, to move away from each other, results in a more diverse set for $u_0$; in this case, without compromising accuracy.\looseness=-1
	}
	\label{fig:illust}
\end{figure*}



\subsection{Strategic content creators}
Our key modeling assumption is that items are owned by strategic \emph{content creators}
(or `suppliers') whose aim is to maximize exposure to their items \citep{hron2022modeling}.
Content owners act to increase their item's score $s(x_j)$,
which is an average over the scores of potentially relevant users: 
\begin{equation}
s(x_j) = \avg \{f(x_j;u_i) \mid i \in U_j\}
\end{equation}
To preserve equity across content creators,
we use spherical averaging over user representations $u_i$ to maintain unit norm\footnote{To see why normalizing $v$ is important, consider an item $j$ with two users: if $u_1,u_2$ are close, then $v_j$ will have a similar norm, but if $u_1,u_2$ are spread out, $\|v_j\|$ can be significantly smaller.}, which pertains to the following form: 
\[
s(x_j) = 
\vtilde_j^\top x_j,
\,\,\text{ where }\,
\vtilde_j = \frac{v_j}{\|v_j\|}, \,
v_j = \frac{1}{m_j} \sum\nolimits_{i \in U_j} u_i 
\]
This can be taken to mean that the system reveals normalized scores,
so that utility for item $j$ derives from $\vtilde_j$
which describes an `average' user
representing all $i \in U_j$.

Following the general formalism of strategic classification \citep{hardt2016strategic}, we assume content creators can modify their item's features, at a cost, and in response to the learned predictive model.
Given a known cost function $c(x,x')$,
content creators modify items via the \emph{best response} mapping:\looseness=-1 
\begin{equation}
\label{eq:best_response}
x^f_j = \Delta_f(x_j) \triangleq \argmax_{x':\|x'\| = 1} \,
s(x')
- \alpha c(x_j, x') 
\end{equation}
where the norm constraint ensures that modified items 
remain on the unit sphere,
and 
throughout we consider quadratic costs,
$c(x,x') = \|x-x'\|_2^2$.\extended{\footnote{As in \citet{levanon2021strategic,levanon2022generalized,jagadeesan2022supply,eilat2022strategic}.}}
The scaling parameter $\alpha \ge 0$ will allow us to vary the intensity of strategic updates:
when $\alpha$ is small, modifications are less restrictive
and so $x^f_j$ can move further away from $x_j$, and vice versa.\looseness=-1

We consider item modification to be `real', in the sense that changing $x$
can cause $y$ to also change.
Following \citet{shavit2019extracting, rosenfeld2020predictions},
we assume labels are determined by an unknown
stochastic ground-truth function $f^*$
which determines personalized relevance for any counterfactual item $x$ as $y=f^*(x;u^*_i)$,
where $\{u^*_i\}_{i=1}^m$ are ground-truth user preferences
(which are unknown to the learner).\looseness=-1
\extended{This can be relaxed to assuming a fixed conditional $p^*(y|x;u_i^*)$.}

\textbf{Strategic behavior and diversity.}\,\, 
Eq. \eqref{eq:best_response} reveals how the system can drive incentives: since each content creator $j$ acts to make their item more aligned with $\vtilde_j$, the system can set the $u_i$
(which together compose all $\vtilde_j$) to induce $\vtilde_j$-s that vary in their orientation;
this incentivizes different content creators to move towards different directions---thus creating diversity,
whose potential growth rate is mediated by $\alpha$.
Note that even linear $f$ can incentive different items to move in different directions, since
(i) due to norm constraints, items will not necessarily move in the direction of the gradient of $f$;
and more importantly, (ii) since different items appeal to different users,
each $\vtilde_j$ defines a utility function $s(x_j)= f(x_j;\vtilde_j)$
that is distinct for item $j$.
Nonetheless, the $\vtilde_j$ are not disjoint;
the connectivity structure in $G$ forms dependencies across the $u_i$,
which introduce correlations in how 
items can jointly move
(see Fig. \ref{fig:illust} (B)).


\extended{
\paragraph{Label updates.}
Contrary to vanilla strategic classification,
we allow changes in item features $x$ to trigger changes
in corresponding relevance scores $y$.
This is in line with works on causal strategic learning,
in which modifications to $x$ can causally affect outcomes
downstream.
For concreteness,
and similarly to \citet{shavit2019extracting, rosenfeld2020predictions},
we assume labels are determined by an unknown
ground-truth function $f^*$
which determines personalized relevance for any counterfactual item $x$ as $y=f^*(x;u^*_i)$,
where $\{u^*_i\}_{i=1}^m$ are ground-truth user preferences
(which are unknown to the learner).
From a modeling perspective,
relying on $f^*$ for determining counterfactual outcomes
is useful since it makes our problem one of \emph{covariate shift} \citep{bickel2009discriminative},
for which retraining is a sound approach.
Note $f^*$ can be quite distinct from $f$ (e.g., can be non-linear), and in our experiments we take several measures
to differentiate them.}


\subsection{Interaction dynamics}
\label{subsec:dynamics}
We will be interested in studying how learning affects ranking accuracy and diversity over time.
As noted, we focus on retraining dynamics,
where at each round $t$ the system re-trains its predictive model
$\smash{f^t}$ on current data $\smash{(x^{t},y^t)}$,
which in our case, is based on strategic responses to the previous model, $\smash{x^{t}_j = \Delta_{f^{t-1}}(x^{t-1}_j)}$.
We think of item modification as a process which takes time:
In the initial portion of round $t$,
users remain to observe $\smash{x_j^t}$, on which $f^{t}$ was trained;
but exposure to $f^{t}$ incentivizes change,
and so after some time has passed,
content creators publish the modified
$\smash{x_j^{t+1}}$,
for which users provide fresh labels $\smash{y_{ij}^{t+1}}$.
Once data has been collected for all modified inputs,
the system retrains.

\extended{
We model interactions between the learning system, strategic content creators, and users as follows.
Interactions occur over time and in discrete rounds,
where each round $t$ includes three steps:
(i) At the beginning of the round,
the system uses data collected in the previous round 
to learn current user representations $\smash{u_i^{(t)}}$,
which it then uses for recommendations $\smash{\r_i^{(t)}}$;
(ii) During the round, content creators respond to the system
by strategically modifying their corresponding items $\smash{x_j^{(t)} \mapsto x_j^{(t+1)}}$;
(iii) Finally, towards the end of the round,
users consume recommended items and report relevance ratings $\smash{y_{ij}^{(t)}}$ for the updated $\smash{x_j^{(t)}}$,
which are then used as input for the system in the 
next round.
}

\extended{
\subsection{Relation to performative prediction}
- performative
  -- retraining
  -- stateful!
- what changes (x and y!), D operator
  -- same population, so no actual `distribution' to change
- time scales
  -- `even' and `odd' times
  -- 1st stage recs (=lists) do not change
- `constraint': want ndcg to be good *now*
- strategic updates take time; want to have good div *later*, before next retraining
- pre/post: definition, meaning

Note that unlike other works in which utility derives from a linear predictive model
\tocitec{strat reg, strat ranking; others?},
in our setting different items move in different manners.
This is since:
(i) items must remain on the unit sphere
(which also implies that cost budgets do not necessarily deplete), and
(ii) utility is personalized per item (due to different user sets $U_j$).
As we will show, the latter point is crucial for creating diversity.
}

%% file: method.tex
\section{Learning and Optimization}
\label{sec:Learning}


The conventional approach for learning to recommend
relies on training predictive models to correctly rank
items by their relevance. 
Then, to promote diversity,
a post-hoc procedure is typically used to re-rank
$\r$,
which in our setting can affect diversity by determining which items appear in the top $k$. 

The main drawback of re-ranking is that its heuristic nature
means that diversifying the list might reduce its relevance,
and can cause NDCG to deteriorate substantially.
As an alternative, here we pursue a more disciplined approach,
in which we directly optimize the joint objective:
\begin{equation}
\label{eq:learning_objective_non-strat}
\max_{u_1,\dots,u_m} \frac{1}{m} \sum_{i=1}^m \NDCGk(\y_i,\r_i) + 
\lambda \, \divk(X_i,\r_i)
\end{equation}
where $\r_i = \rank(f(X_i;u_i))$ is the ranking of items in $X_i$ for user $i$,
and $\lambda$ trades off between accuracy and diversity.
With $\diversity$ as regularization,
we can tune $\lambda$ to obtain a desired balance,
or maximize diversity under accuracy constraints.
In our experiments we tune $\lambda$ to achieve a predetermined level of NDCG (e.g., 0.9); in this case, regularization serves as a criterion for choosing the most diverse model out of all sufficiently-accurate models.
%
We next describe our approach for optimizing
Eq. \eqref{eq:learning_objective_non-strat},
which sets the ground for our strategically-aware objective.
\looseness=-1

\subsection{Optimization} \label{sec:optimization}
We propose to optimize Eq. \eqref{eq:learning_objective_non-strat}
by constructing a differentiable proxy objective,
to which we can then apply gradient methods.
The key challenge is that Eq. \eqref{eq:learning_objective_non-strat} relies on a ranking operator (i.e., for computing NDCG and top-$k$),
which is non-differentiable.
Our approach adopts and extends \citet{Pobrotyn2021NeuralNDCGDO},
and makes use of the differentiable sorting operator introduced in \citet{grover2018stochastic}.
First, consider NDCG.
For the numerator, note that the ranking operator $\r(\cdot)$ can be implemented
using a corresponding permutation matrix $P$, i.e., 
$y_{\r(\ell)} = (Py)_\ell$. 
To differentiate through $\r$, we replace $P$ with a `smooth'
row-wise softmax permutation matrix,
$\Phat$. 
We denote by $\rhat$ the corresponding soft ranking,
computed as 
$\rhat = (\Phat \odot Q)^\top \bm{1}$,
where $Q_{ij} = i \,\,\forall j$, $\odot$ is the Hadamard product,
and $\bm{1}$ is a vector of 1-s.
The denominator for $\NDCGk$ requires accessing indexes in $y$ using explicit entries in $\r$;
applying $\rhat$ instead gives a weighted combination of $y$-s, with most mass concentrated at the correct $\r(\ell)$ when $\rhat$ is a good approximation.
The summation term is implemented using a soft top-$k$ operator,
which we obtain by applying an element-wise scalar sigmoid to $\r$ as:
\begin{equation}
\1{r \le k} \approx \sigma_{\tau}(k-\rhat)    
\end{equation}
For the diversity term, note that $\cos$ is naturally differentiable.
To make the entire $\divk$ differentiable,
we use the soft top-$k$ operator on item pairs via:
\begin{equation}
\label{eq:differentiable_div}
\frac{1}{k(k-1)} \sum_{j,j'}^n
\sigma_\tau(k-\rhat_j) \sigma_\tau(k-\rhat_{j'})
\left(1 - x_j^\top x_{j'}\right)
\end{equation}



\subsection{Strategically-aware learning}
Although Eq.~\eqref{eq:learning_objective_non-strat} accounts for diversity in learning, it does so \emph{reactively}, in a way that is tailored to the previous time step.
Since the learned $f$ incentivizes content creators to modify items,
we propose to promote diversity \emph{proactively} by anticipating their strategic responses.
To do this, we replace $X_i$ with the anticipated
$X^f_i =  \Delta_f(X_i) = \{\Delta_f(x)\}_{x \in X_i}$
to get our strategically-aware objective:
\begin{equation}
\label{eq:learning_objective_strat}
\max_{u_1,\dots,u_m} \frac{1}{m} \sum_{i=1}^m \NDCGk(\y_i,\r_i) + 
\lambda \, \divk(X^f_i,\r^f_i) 
\end{equation}
where $\r^f_i = \rank(f(X^f_i;u_i))$ is the anticipated ranking.
Eq.~\eqref{eq:learning_objective_strat} optimizes for pre-modification NDCG,
but promotes post-modification diversity; this is the mechanism through which
learning can incentivize the creation of diversity.\footnote{While in principle it may be possible to also consider future NDCG,
note this necessitates reasoning about how deploying $f$ affects future $y$, which is a challenging causal inference task.}\looseness=-1

\extended{with $\diversity$ now serving as a `lookahead' regularizer
that anticipates outcomes in the next time step.}

The challenge in optimizing Eq.~\eqref{eq:learning_objective_strat} is
twofold: (i) $\Delta_f$ is an argmax operator, which can be non-differentiable,
and (ii) $\Delta_f$ depends on $f$ both internally (by determining utility)
and externally (in the top-$k$ operator of $\divk$).
Fortunately, for our modeling choices,
$\Delta_f$ can be solved in a differentiable closed form.
Using KKT conditions, we can derive:
\begin{equation} \label{eq:Delta_closed_form}
\Delta_f(x_j) = \frac{\vtilde_j + 2\alpha x_j}{|\vtilde_j + 2\alpha x_j|}
\end{equation}
Proof in Appendix \ref{apx:best_response}.
Plugging Eq.~\eqref{eq:Delta_closed_form}
into Eq.~\eqref{eq:learning_objective_strat}
gives us our final differentiable strategic learning objective.

%% file: analysis.tex
\section{Diversity via Strategic Incentives}

\label{sec:diversity_strategic_incentives}
The reliance of the utility of content creators on the learned $f$
provides the system with potential power for shaping incentives.
Here we analyze when this potential can materialize
in a simplified setting 
that focuses exclusively on maximizing diversity,
for a single time step with no cost restrictions (i.e., set $\alpha=0$).
This removes any constraints on $u_i$ that may arise from accuracy considerations,
and serves as a convenient substitute for lengthy strategic dynamics. 

\extended{In Appendix \ref{apx:empirical_generalization_full_overlapping} we empirically show that this is indeed representative of $\alpha>0$ over time (absent retraining).}

Our main object of interest in the analysis is the recommendation graph $G$,
viewed as input to the learning algorithm. 
As we show, whether diversification through incentivization is possible (or not)
depends on properties of the graph.
We first consider the graph of a single item list $X$ and all related users (i.e., users $u_i$ for which some $x_j \in X_i$ is also in $X$),
and then proceed to general graphs over multiple lists.

\begin{figure*}[t!]
\centering
\includegraphics[width=0.33\textwidth]{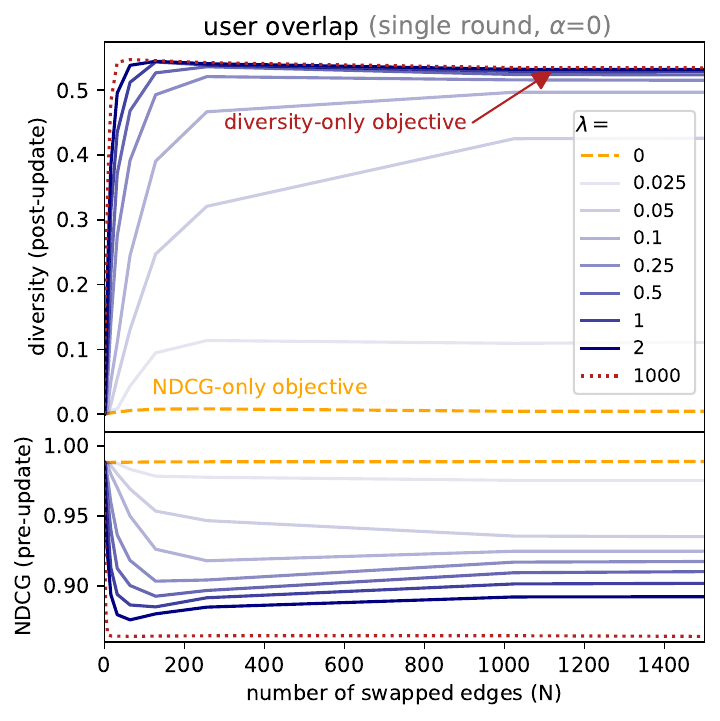}
\includegraphics[width=0.33\textwidth]{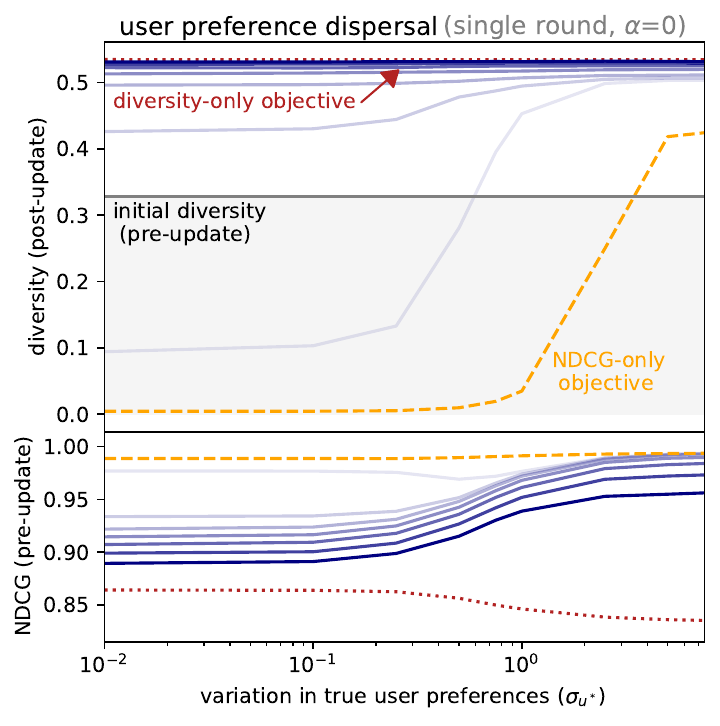} 
\includegraphics[width=0.33\textwidth]{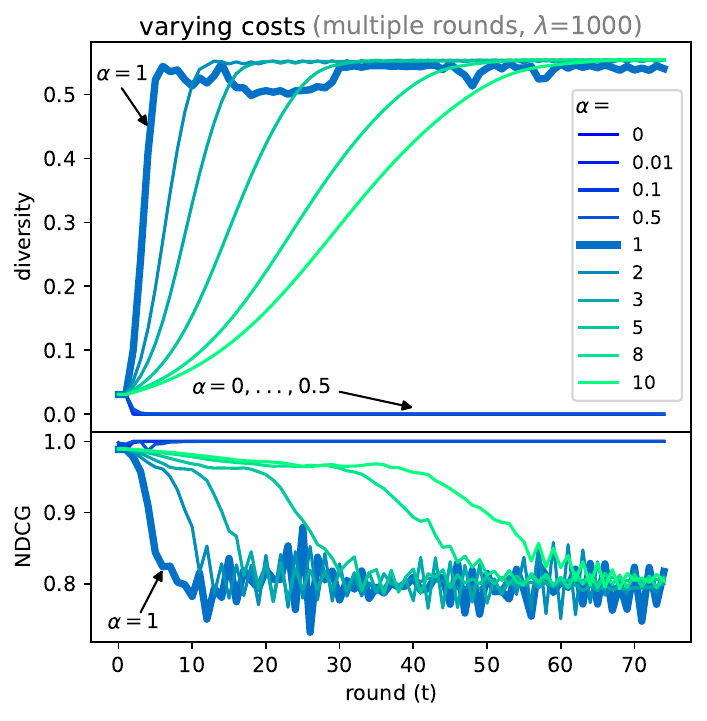} 
\caption{
\textbf{Synthetic experiments.}
\textbf{(Left)} 
As user overlap decreases (larger $N$), our approach is quickly able to incentivize diversity; this is pronounced for larger $\lambda$.
\textbf{(Center)} 
For any dispersion level of true user preferences ($u^*$),
even mild regularization ($\lambda>0$) suffices for our approach to improve diversity; without it ($\lambda=0$), diversity may plummet.
\textbf{(Right)} 
For low cost scales ($\alpha<1$), changes are large,
and learning is unable to diversify.
But once $\alpha \ge 1$, this changes sharply,
with Larger $\alpha$ requiring more time for diversity to 
smoothly form. 
}
\label{fig:synth}
\end{figure*}

We begin with a negative result for single lists. 
\begin{proposition}
\label{prop:Full_users_overlap}
Let $X=\{x_1,x_2\}$.
If both items have
fully overlapping users
(i.e., $U_1 = U_2$), then for any $f$, 
$\diversity(X^f,\r^f)=0$.
Hence, diversity cannot be incentivized.
\end{proposition}
Proof in Appendix \ref{apx:proof_full_users_overlap}.
Prop. \ref{prop:Full_users_overlap} shows how similar users induce similar incentives, resulting in $\Delta_f(x_1) = \Delta_f(x_2)$.
Extending the result to larger item sets and multiple users is straightforward, and implies the following:
if the same list of potentially-relevant items is associated
exclusively with the same group of users, 
then strategic behavior is bound to nullify diversity entirely---regardless of any system efforts.
%
Conversely, Prop. \ref{prop:Full_users_overlap} hints that to diversify items,
it is necessary to start out with some initial variation in the assignment of users to items. But how much user variation is needed?
Our next result shows that for a single item list, when no other considerations are present,
minimal differentiation in users is sufficient for obtaining maximal item diversity.
\begin{proposition}
\label{prop:diff_users_max_div}
Let $X=\{x_1,x_2\}$.
If $U_1,U_2$ differ only in a single user,
i.e., if $U_1 = U\cup\{u_1\}$ and $U_2 = U\cup\{u_2\}$ for some $U,u_1,u_2$ with $|U|>1$,
then there exists an $f$
which obtains maximal diversity,
i.e., $\diversity(X^f,\r^f)=1$.
\end{proposition}  
The proof is constructive, and appears in Appendix \ref{apx:proof_diff_users_max_div}.
Prop. \ref{prop:diff_users_max_div} shows that,
under lenient conditions,
incentivization can drive diversity to its greatest possible extend.
Generally, and under more realistic considerations (e.g., accuracy and cost constraints),
we expect that greater differentiation is likely necessary,
even for lower gains in diversity.

We now  move to considering graphs for 
multiple item lists.
This introduces dependencies:
if some item $x_j$ appears in two distinct but partially-overlapping user lists $X_i,X_{i'}$,
then this restricts the possible values that the embeddings $u_i,u_{i'}$ can take.
As such, Prop. \ref{prop:diff_users_max_div} cannot simply be applied
to each list in $G$ independently, since $f$ cannot be tailored to maximize diversity for $X_i$ without affecting $X_{i'}$, for which $f$ may not be optimal.
Our final result shows that, despite such dependencies,
significant diversity is still attainable.


\green{
}
\begin{proposition}
\label{prop:div_multi_list}
Let $K=2$, then 
for any $\epsilon>0$ and any $N$,
there exists a graph $G$ over $N$ distinct lists
and a corresponding $f$ s.t. the average diversity is at least
$(1-\epsilon)(1-3/N)$.
\end{proposition}
Proof in Appendix \ref{apx:proof_div_multi_list}, and relies on a construction that
simultaneously (i) decouples users across lists, and
(ii) permits maximal diversity within lists.
Prop. \ref{prop:div_multi_list} shows that
the potential for incentivizing for diversity grows towards the near-optimal value of $1-\epsilon$ quickly,
in terms of the number of lists $N$---and to the degree that the graph permits.

%% file: synthetic.tex
\section{Synthetic Experiments}
\label{sec:synthetic}
Our previous section showed that,
under favorable conditions,
it is theoretically possible 
to generate significant diversity through incentivization.
In this section we empirically demonstrate,
in a series of increasingly-complex synthetic tasks,
how our strategically-aware learning approach (Eq.~\eqref{eq:learning_objective_strat}) can encourage
diversification in practice.
Each task is designed to explore a different factor in our setup,
and to shed light on how accuracy and diversity trade off as $\lambda$ is varied.
Appendix~\ref{apx:additional_result} includes additional results.\looseness=-1


\textbf{Experimental setup.}\,\,
We set $n=200,m=50,k=K_i=10$ for all $i$,
and fix $d=2$ so that features $x,u$ can be easily visualized as angles.
Item features $x$ are sampled from
$\mathcal{N}((\sfrac{1}{\sqrt{2}},\sfrac{1}{\sqrt{2}}),\sigma^2_x I)$.
We use $f^*(x;u_i^*) = 2^{(u_i^*)^\top x}$,
where ground truth user preferences $u^*$
are sampled from
$\mathcal{N}((\sfrac{1}{\sqrt{2}},\sfrac{1}{\sqrt{2}}),\sigma^2_{u^*} I)$.
All features are normalized post-sampling.
We use $\sigma_x=1$ and $\sigma_{u^*}=0.1$,
but in some settings vary them to control dispersion.
All results are averaged over 100 random repetitions (when applicable).



\subsection{The role of variation in user item lists}
Our first experiment investigates the importance of variation across user item lists,
which complements Sec.~\ref{sec:diversity_strategic_incentives}.
We begin with a graph composed of five mutually-exclusive fully-connected subgraphs, exhibiting full overlap;
then, we `shuffle' $N$ edges across subgraphs, for increasing $N \in \mathbb{N}$---which decreases user overlap (at $N \approx 1000$ edges are approx. uniform).
%
Figure \ref{fig:synth} (left) shows NDCG
(bottom)
and diversity
(top)
for a range of $\lambda$.
For $N=0$ (full overlap), diversity is zero for all $\lambda$,
in line with Prop. \ref{prop:Full_users_overlap}.
As $N$ grows, diversity increases, but at the cost of reduced NDCG, which is more pronounced for larger $\lambda$.
Note how only minimal overlap (e.g., $N=100$ for $\lambda=1$) suffices for generating considerable diversity, which rises sharply once $N>0$.
This suggests our approach can utilize the capacity for diversity implied by Prop. \ref{prop:diff_users_max_div},
even under accuracy constraints.\looseness=-1


\begin{figure}[b!]
\centering
\includegraphics[width=\columnwidth]{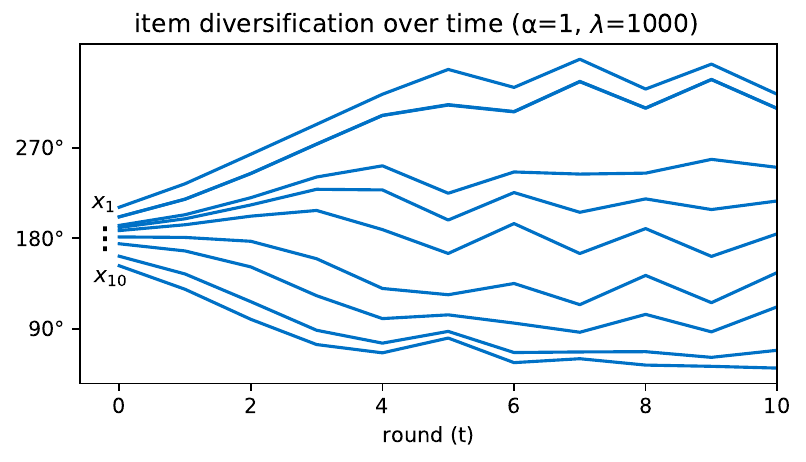}
\caption{
Dispersal of items $x_i \in \R^2$ over time.
}
\label{fig:angles}
\end{figure}

\subsection{The role of variation in true user preferences}
\label{variation_u*}
When learning aims primarily for accuracy,
training encourages each $u_i$ to be oriented towards its $u^*_i$.
For diversity, this acts as a constraint
which restricts the capacity of $f$ to diversify.
Here we study the role of variation in $u^*_i$ as a mediator in this process.
Figure \ref{fig:synth} (center) shows NDCG
(bottom)
and diversity
(top),
both in absolute values and relative to pre-update diversity,
for varying $\sigma_{u^*}$ and for a range of $\lambda$.
Here we sample edges uniformly, and consider a single time step with $\alpha=0$.
Without regularization ($\lambda=0$),
strategic updates cause diversity to drop to zero,
even when user preferences are reasonably dispersed ($\sigma_{u^*}=1$).
However, even mild regularization ($\lambda=0.05$) suffices for obtaining significant diversity through incentivization,
which becomes more pronounced as $\lambda$ grows;
for all $\lambda>0.1$, diversity is high, and in effect remains fixed.
This suggests our model can effectively utilize natural variation in user preferences.
Increased diversity comes at the cost of NDCG, 
but this diminishes quickly for larger dispersion.

%



\begin{figure*}[t!]
    \centering
    \includegraphics[width=0.9\textwidth]{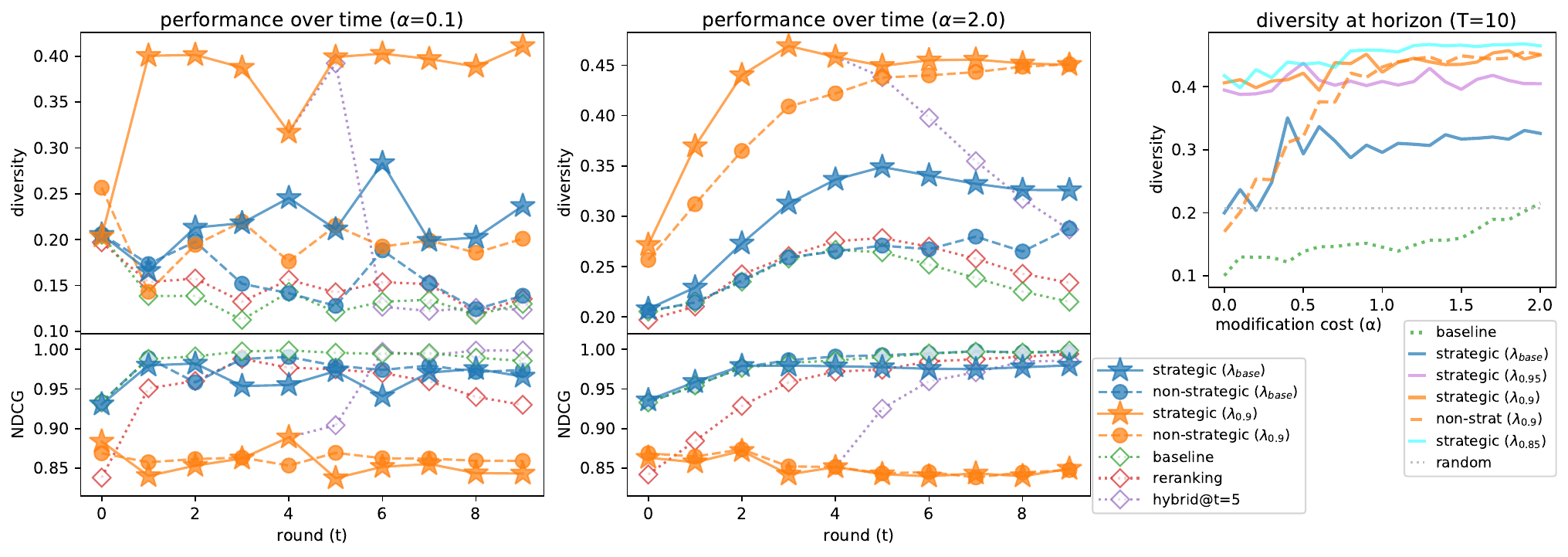}
    \caption{
    \textbf{Experiments on Yelp data.}
    \textbf{(Left+center)} Diversity and NDCG over rounds for different methods and target NDCG values,
    for loose ($\alpha=0.1$; left)
    vs. restricted ($\alpha=2.0$; center) strategic updates.
    Our \method{strategic} approach consistently achieves significant and sustained improvement in diversity, with minimal loss in accuracy.
    \textbf{Right} Diversity after $T=10$ rounds
    for increasing costs scales $\alpha$.
    }
    \label{fig:experiements_real}
\end{figure*}

\subsection{The role of time vs. modification costs} \label{sec:synth_vary_costs}
We now turn to examining the temporal formation of diversity
through retraining dynamics, as mediated by modification costs $\alpha$.
When $\alpha$ is large, content creators can only apply small changes to $x$ at each round (and vice versa for small $\alpha$).
On the one hand, small steps suggest that diversity may require time to form; but on the other, note that small steps also allow the system to intervene with high fidelity and direct incentives throughout, and hence to gradually `steer' behavior towards diversification.
%
%
Figure \ref{fig:synth} (right) shows
diversity (top) and NDCG (bottom) 
for increasing $\alpha$ and over multiple retraining rounds.
We set $\lambda$ to be large
so that learning is geared primarily towards diversity.
When $\alpha$ is small (here, $<1$), diversity quickly drops to zero---as in Prop. \ref{prop:Full_users_overlap}.
In contrast, $\alpha=1$ exhibits a sharp transition,
in which diversity quickly rises. 
Fig. \ref{fig:angles} visualizes for a set of items in $\R^2$ how they quickly become dispersed.
Larger $\alpha$ entail similarly high diversity:
here the process is slower---since update steps are smaller,
but also more stable---since the system has finer control over each step;
c.f. $\alpha=1$, where NDCG fluctuates. 

%% file: real_data_exp.tex
\section{Experiments on Real Data}
\label{sec:real data}
We now turn to evaluating our approach on real data.
Here we study how NDCG and diversity evolve over time under different
learning methods and experimental conditions.
See Appendix~\ref{apx:Data details} for additional details,
and Appendix~\ref{apx:results_real} for extended results (\ref{apx:additional_div},\ref{apx:additional_tradeoff}),
a sensitivity analysis to misspecification of $\alpha$ (\ref{apx:sensitivity}),
and additional similarity metrics (\ref{apx:entropy}).\looseness=-1


\paragraph{Data.}
Our experimental setup is based on the restaurants portion of the Yelp dataset\footnote{\url{https://www.yelp.com/dataset/download}},
which includes user-submitted restaurant reviews.
We focus on users having at least 100 reviewed restaurants.
For each user $i$, we construct the list of potential items $X_i$
to include the 40 most popular restaurants of those reviewed by $i$,
which amounts to 236 users and 1,520 restaurants in total.
We elicit $d=43$ restaurant features
(e.g., cuisine type, noise level)
to be used by the system for learning.
We also elicit `ground truth' user features $u^*_i$
used for optimizing $f^*(x',u^*_i)$,
which is trained to predict the likelihood that $i$ will review $x$,
interpreted here as relevance $y$.
The labeling function $f^*$ is used only for determining
updated relevancies $y'$ for modified items $x'$;
neither $f^*$ nor the $u^*_i$ are known to the learner.
We ensure $f^*$ is distinct from learnable functions $f$ by several means:
(i) $f^*$ is a fully-connected deep network, whereas $f$ are linear;
(ii) $f^*$ is trained using true user features $u^*_i$, which are unobserved for $f$;
(iii) $f^*$ is trained on considerably more data,
and of which the data used for training $f$ is a non-representative subset;
and
(iv) $f$ is trained on labels that are modified to emphasize highly-rated items.
Full details in Appendix \ref{apx:training f^*}.\looseness=-1

\paragraph{Setup and learning.}
We consider top-10 recommendation (i.e., $k=10$),
and evaluate performance over $T=10$ rounds of retraining and corresponding strategic updates.
For training, 
we assume that at each round the system has access to 30 items per user,
randomly selected (out of the 40) per round;
of these, a random 20 are used for training, and the remaining 10 are added for validation
(tuning $\lambda$ and early stopping).
Test performance is evaluated on all 40 items.
Since the test set includes additional items (compared to training),
$f$ must learn to generalize well to new content at each step.
Since adding items also changes the graph,
and since the graph determines strategic responses---$f$ must also learn to generalize to new forms of strategic updates.

\paragraph{Methods and evaluation.}
In line with our dynamic setup (Sec.~\ref{subsec:dynamics}),
all methods considered aim primarily at optimizing `current' NDCG
(i.e., at time $t$ maximize NDCG on $x^{(t)}$),
but differ in how (and if) they promote diversity.
These include:
(i) a \method{non-strategic} approach which regularizes for current diversity on non-strategic inputs $x^{(t)}$ (Eq. \eqref{eq:learning_objective_non-strat}),
(ii) our \method{strategic} approach, which regularizes for future diversity on the anticipated $x^{(t+1)} = x^{f_t}$ (Eq. \eqref{eq:learning_objective_strat}),
(iii) an accuracy-only \method{baseline}, which does not promote diversity
(by setting $\lambda = 0$),
(iv) \method{re-ranking} using the popular \method{MMR} diversification procedure \citep{carbonell1998use},
and (v)
a \method{hybrid@$t$} approach, which runs \method{strategic} for $t=5$ rounds, and then `turns off' regularization.\looseness=-1

Since \method{non-strategic} and \method{strategic} are designed to balance NDCG and diversity,
for a meaningful comparison,
in each experimental condition we fix a predetermined target value for NDCG,
tune each method at each round to achieve this target
(using $\lambda$, on the validation set,
and up to tolerance 0.01),
and compare the resulting diversity.
We use the notation $\lambda_{\beta}$ to mean that $\lambda$
was tuned for the target $\beta$.
To compare with \method{baseline}, in one condition we set $\lambda$ to the largest value that maintains the same NDCG as the baseline (for which $\lambda=0$),
denoted $\lambda_{\text{base}}$.\looseness=-1
%
Note that due to performativity, data at time $t$ depends on the learned model at time $t-1$. Because our dynamics are \emph{stateful}, results are path-dependent,
and so comparisons must be made across full trajectories---and cannot be made independently at each time point.

\subsection{Diversity over time}
\label{sec:div_over_time}
Our first experiment studies how diversity evolves over time
for different NDCG targets.
Fig.~\ref{fig:experiements_real} (left) shows diversity (top) and NDCG (bottom) per round
for $\alpha=0.1$, which permits significant (yet restricted) strategic updates.
Results show that for \method{baseline}, which does not actively promote diversity,
average diversity decreases over time by roughly 40\%.
Adding \method{MMR}, which re-ranks for diversity, improves only marginally.
In the $\lambda_\text{base}$ condition,
diversity for \method{non-strategic} improves only slightly
compared to \method{baseline}, mostly at the onset.
In contrast, our \method{strategic} approach is able to consistently maintain (and even slightly increase) diversity over time,
albeit at small occasional drops in test NDCG, as compared to \method{non-strategic} ($\le 0.02$).\footnote{One possible reason for \method{baseline} to achieve NDCG$\approx$1 is that when diversity is extremely low,
items are so similar (and hence similarly relevant) that choosing the top-$k$ becomes trivial.}
In the $\lambda_{0.9}$ condition, \method{non-strategic} is able to preserve roughly 80\% of diversity by sacrificing $\sim 13\%$ of the optimal NDCG.
Meanwhile, and for the same loss in NDCG,
\method{strategic} is able to quickly \emph{double} the initial diversity---and sustain it, likely through means similar to those observed in Sec. \ref{sec:synth_vary_costs}.
Note that sustaining diversity requires to actively promote it throughout; once regularization is switched off (\method{hybrid}),
diversity immediately drops.

Fig.~\ref{fig:experiements_real} (center) shows results for a similar setting, but using a larger $\alpha=2$.
In comparison, here diversity for all methods improves---but for \method{baseline} and \method{MMR}, this is only transient.
Results are also smoother, and accumulate---for both improvement and deterioration---which is likely due to the fact that strategic movements are now more restricted.
The biggest distinction here is that for $\lambda_{0.9}$, \method{non-strategic} eventually obtains the same level of diversity as \method{strategic},
this likely due to $\alpha$ being large:
small updates mean that $x$ and $x^f$ are correlated,
and so regularizing for $\diversity(X,\r)$
also improves $\diversity(X^f,\r^f)$ to some extent.
Here the advantage of 
\method{strategic} is that it improves much faster.


\begin{figure}[t!]
    \centering
    \includegraphics[width=\columnwidth]{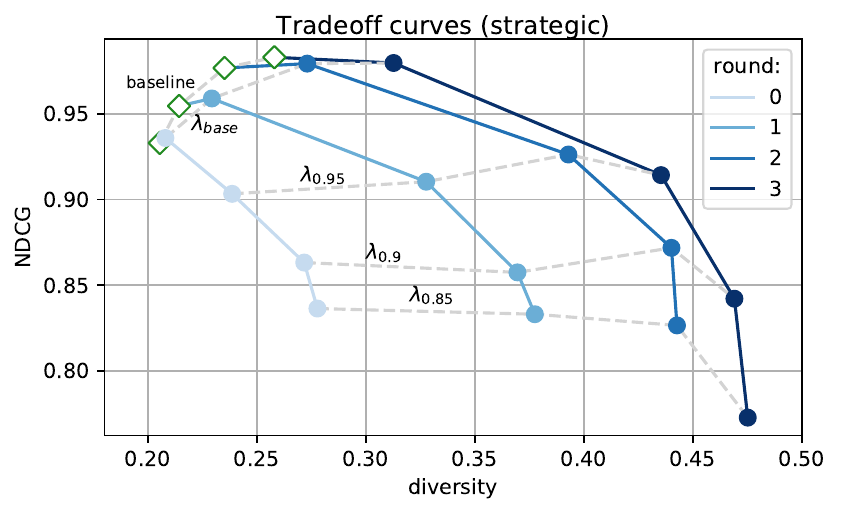}
    \caption{
   Pareto curves for \method{strategic},
    per round (solid lines) and per $\lambda$ (dashed).\looseness=-1
    }
    \label{fig:tradeoff}
\end{figure}

\subsection{The role of modification costs.}
\label{sec:tradeoff}
To further examine the role of $\alpha$,
we compare performance across a range of cost scales
and for multiple target NDCG values.
Fig. \ref{fig:experiements_real} (right) shows
diversity at $T=10$.
As can be seen, all methods benefit from increasing $\alpha$.
However, diversity for \method{baseline} remains below 
that of a \method{random} baseline which recommends $k$ random items.
For \method{strategic}, sacrificing fairly little NDCG ($\lambda_{0.95}$ and below) suffices for gaining significant diversity across \emph{all} $\alpha$.
Results also show how $\alpha$ mediates the gap between \method{non-strategic} and \method{strategic}, which increases as $\alpha$ grows.\footnote{To avoid clutter we plot \method{non-strategic} only for $\lambda_{0.9}$, which is comparable to Fig. \ref{fig:experiements_real} (left) and (center), but note that other target NDCG values exhibit qualitatively similar patterns.}\looseness=-1

\subsection{Tradeoffs over time.}
Fig.~\ref{fig:tradeoff} 
shows Pareto curves for \method{strategic},
obtained by considering multiple $\lambda$
and over several rounds.
Results show that the entire tradeoff curve between NDCG and diversity improves over time.
Paths for each target NDCG depict specific trajectories,
and show how diversity can be created without sacrificing accuracy.
This holds even when no compromise in accuracy is allowed:
even though both \method{baseline} and $\lambda_{\text{base}}$ maximize accuracy,
our \method{strategic} approach is able to steer content creators towards more diverse content.
This is achieved by utilizing the flexibility to choose $u_i$ which improve both accuracy \emph{and} diversity.

\extended{
\todo{add pre/post update results}

Recall that our dynamic setup (Sec. \toref) assumes that updates
occur over time.
At each step $t$ we report the performance
on the current data $x^{(t)}$
of the previous model $f^{(t-1)}$ (trained on $x^{(t-1)}$) and the current-trained model $f^{(t)}$ (trained on $x^{(t)}$).

Since we model strategic updates as taking place over time,
intuitively we would expect that performance for $f^{(t-1)}$ degrade from $t-1$ to $t$, but due to accuracy constraints, that all post-training accuracies will be the same. This lets us focus on how diversity evolves.

- when retraining, points don't move; only reason div changes is because top-k set can change (otherwise, div would be the same!) \\
- note retraining for lookahead improves div; in principle, objective was supposed to optimize div before retraining (and after movement); but retraining can help if $x'$ and $x''$ are correlated \\

}

%% file: Discussion.tex
\section{Discussion}
\label{sec:discussion}

Through the lens of performativity,
our paper studies how a learning system can incentivize content creators to collectively form a more diverse inventory of items for recommendation.
In this, we challenge the conventional view that diversity is merely a matter of which items to present (and which not),
and argue that to fundamentally rectify the predisposition of modern recommendation systems to homogenize content,
learning must (i) recognize that content \emph{changes} over time,
in part due to the strategic behavior of content creators, and
(ii) capitalize on its power shape incentives and steer towards diversity.
Our work joins others in taking a step towards studying recommendation systems as complex ecosystems in which users, creators, and the system itself act and react to promote their own goals and aspirations.
From a learning perspective, this requires us to rethink the role that predictions play in the recommendation process,
and consider its implications on user welfare.

\todo{limitations; future work}


%% file: appendix.tex
\section{Optimization}
\subsection{Closed form expression for best response $\Delta_f$} \label{apx:best_response}
\proof
To compute an item's best-response update $\Delta_f(x)$,
we must solve the following constrained problem:\\
\begin{align*}
& \max_{x'} v^Tx' - \alpha c(x', x) \quad \textrm{s.t} \quad ||x||_2=1 &   \\
\intertext{where $c(x,x')=\|x'-x\|_2^2$.
We solve for $x'$ using Lagrangian analysis.
First, we square the constraint (which does not change the condition itself). 
Define the Lagrangian as follows:}
&L(x', \lambda) =  v^Tx' - \alpha\|x' - x\|_2^2 +\lambda[||x'||_2^2-1] &
\intertext{Next, to find the minimum of $L$, derive with respect to $x'$, and compare to 0:}
& v - 2\alpha(x' - x) + 2\lambda x' = 0  \\
& v + 2\alpha x = 2\alpha x' - 2\lambda x' \\
& \frac{v + 2\alpha x}{2\alpha - 2\lambda} = x' &
\intertext{Plugging $x'$ into the original constraint gives:}
& \frac{|v+2\alpha x|}{|2\alpha - 2\lambda|} = 1 \\
& |v+2\alpha x| = 2\alpha -2\lambda \\
& |v+2\alpha x| - 2\alpha = -2\lambda & \\
\intertext{Finally, plugging $\lambda$ into the expression for $x'_i$ obtains:}
& \frac{v +2\alpha x}{|v+2\alpha x|} = x'
\end{align*}

\section{Proofs}
\subsection{Proposition \ref{prop:Full_users_overlap}}
\label{apx:proof_full_users_overlap}
\proof
As both items have the same set of users, the normalized average of user embeddings is the same for both items, i.e., $\vtilde_1=\vtilde_2=\vtilde$. Since $\alpha=0$, the best response (Eq. \eqref{eq:Delta_closed_form}) is given by $\Delta_f(x_j) = \vtilde$ for both $j=1,2$. 
Hence, both items are modified in the same way, and so diversity is by definition zero.
\endproof

\subsection{Proposition \ref{prop:diff_users_max_div}}
\label{apx:proof_diff_users_max_div}
We begin with a useful lemma.

\begin{lemma}
    \label{lem:sum_to_0}
    For any $n>1$ there exists a set of $n$ unit-norm user embeddings $U=\{u_1,\dots,u_n\}$ for which:
    \begin{align*}
        \sum_{u_i \in U} u_i = \overrightarrow{0}
    \end{align*}
    where $\overrightarrow{0}$ is the zero vector.
\end{lemma}

\proof
    We prove for $d=2$; for $d >2$, we set the first two coordinates accordingly, and the other $d-2$ coordinates are set to zero.
    For $n=2$, we define $U_2 = \{u_1, u_2\}$ as follows:
    \begin{align*}
        u_1 = (-1, 0), \quad
        u_2 = (1, 0)
    \end{align*}
    Note that indeed $\sum_{u_i \in U_2} u_i = u_1 + u_2 = (0, 0)$. 

    For $n=3$, we define $U_3 = \{u_1, u_2, u_3\}$ as follows:
    \begin{align*}
        u_1 = (\frac{1}{2}, \frac{\sqrt{3}}{2}), \quad
        u_2 = (\frac{1}{2}, -\frac{\sqrt{3}}{2}), \quad
        u_3 = (-1, 0)
    \end{align*}
    Here as well $\sum_{u_i \in U_3} u_i = u_1 + u_2 + u_3 = (0, 0)$.
    
    \textbf{Observation:} 
    $\forall n > 1$ $\exists a,b \ge 0$ for which
    $n = 2a + 3b$.

    Thus, for any $n > 1$, we define $U$ to include $a$ copies of $U_2$ and $b$ copies of $U_3$.
    Using this construction, we get 
    $\sum_{u_i \in U} u_i = (0, 0)$, as required.
\endproof

We now return to proving the proposition. The general idea is to define two embeddings that are maximally distinct, and
ensure all others do not interfere with their orientation.
\proof
First we prove that for $|U|>1$ diversity is one.
According to Lemma \ref{lem:sum_to_0}, there exists a set of user embedding vectors $U$ of size $|U_1 \bigcap U_2|$ that holds:
\begin{align*}
    \sum_{u\in U} u = \overrightarrow{0}
\end{align*}
We define an $f$ that assigns to each user in $|U_1 \bigcap U_2|$ a vector in $U$ (one to one map).
Since the sum of the vectors in $U$ is 0 it holds that
$\vtilde_1 = u_1$ and $\vtilde_2 = u_2$.
Since $\alpha=0$, there are no modification costs,
and the strategic response of items according to Eq. \eqref{eq:Delta_closed_form} is $\Delta_f(x_1) = u_1$ and $\Delta_f(x_2) = u_2$.
In order to create maximal diversity between $\Delta_f(x_1)$ and  $\Delta_f(x_2)$, we can use any choice of vectors $u_1$ and $u_2$ that satisfy $u_1 = -u_2$.\\
\endproof

\subsection{Proposition \ref{prop:div_multi_list}}
\label{apx:proof_div_multi_list}
We prove for $d=2$; for $d >2$, we set the first two coordinates accordingly, and the other $d-2$ coordinates are set to zero. This allows us to work with angles between user embeddings $u$.
We use `x' and `y' to refer to the corresponding Cartesian components of angles.

Consider the following graph and user vector embeddings.
For each $i < N$, define user $i$'s list of candidate items $X_i$ to include the items $x_i, x_{i+1}$. For user $N$, the items in $X_N$ are $x_N, x_1$.

Let $0<\delta$.  Define the following embedding: 
For every odd $i < N$,
set $u_i = 90 - i\delta$;
For every even $i < N$, set $u_i = 270 - i\delta$.
For $i=N$ we define $u_N = 270 + \delta$.
Note that this gives $x_1=(1, 0)$ since the two users that influence $x_1$ are $u_1$ and $u_N$, whose average gives a vector with coordinates:
\begin{align*}
\text{x}: \quad & 
\frac{1}{2}(cos(270+\delta)+cos(90-\delta))= \\
& \frac{1}{2}(cos(-90+\delta)+cos(90-\delta))= \\
& \frac{1}{2}(cos(-(90-\delta)+cos(90-\delta))= \\
& \frac{1}{2}(cos((90-\delta)+cos(90-\delta))= \\
& cos(90-\delta)\\ 
\text{y}: \quad & \frac{1}{2}(sin(270+\delta)+sin(90-\delta))= \\
& \frac{1}{2}(sin(-90+\delta)+sin(90-\delta))= \\
& \frac{1}{2}(-sin(90-\delta)+sin(90-\delta))= \\
& = 0
\end{align*}
This item's y value is 0; hence, do due unit norm constraints,  we have $x_1 = (1,0)$.

We now calculate the response of $x_i$ when $i$ is even for all $1<i<N$. The users associated with $x_i$ are:
\begin{align*}
u_{i-1} = 90 - (i-1)\delta, \qquad
u_i = 270 - i\delta 
\end{align*}
We average the two vectors by averaging each coordinate $x,y$:
\begin{align*}
\text{x}: \quad &  \frac{1}{2}(cos(270 - i\delta) + cos(90-(i - 1)\delta))= \\
& cos(\frac{360-(2i-1)\delta}{2})cos(\frac{180-\delta}{2}) =  \\ 
& cos(180 - \frac{2i-1}{2}\delta)cos(90-\frac{\delta}{2})= \\ 
& -cos(\frac{2i-1}{2}\delta)sin(\frac{\delta}{2}) \\
\text{y}: \quad & \frac{1}{2}(sin(270-i\delta)+sin(90-(i-1)\delta)= \\
& sin(\frac{360-(2i-1)\delta}{2})cos(\frac{180-\delta}{2})= \\ 
& sin(180-\frac{2i-1}{2}\delta)cos(90-\frac{\delta}{2})= \\
& sin(\frac{2i-1}{2}\delta)sin(\frac{\delta}{2}) 
\end{align*}
Normalizing gives:
\begin{align*}
    \sqrt{(-cos(\frac{2i-1}{2}\delta)sin(\frac{\delta}{2}))^2 + (sin(\frac{2i-1}{2}\delta)sin(\frac{\delta}{2}))^2} =
    sin(\frac{\delta}{2})\sqrt{cos(\frac{2i-1}{2}\delta)^2 + sin(\frac{2i-1}{2}\delta)^2} =
    sin(\frac{\delta}{2})
\end{align*}
Note this must be positive since it describes a vector length; since $0<\delta<180$, this length is $sin(\frac{\delta}{2})$ (and not $-sin(\frac{\delta}{2})$).
The normalized vector is:
\begin{align*}
(-cos(\frac{2i-1}{2}\delta), sin(\frac{2i-1}{2}\delta)) = 
(cos(180-\frac{2i-1}{2}\delta), sin(180-\frac{2i-1}{2}\delta))
\end{align*}
Since $\alpha=0$, according to Eq. \eqref{eq:Delta_closed_form},this gives $\Delta(x_i)$. 

Next, we calculate the response of $x_i$ when $i$ is odd for all $1<i<N$.  The users associated with $x_i$ are:
\begin{align*}
& u_{i-1} = 270 - (i-1)\delta \\
& u_i = 90 - i\delta 
\end{align*}
We average the two vectors by averaging each coordinate $x,y$:
\begin{align*}
\text{x}: \quad &  \frac{1}{2}(cos(270 - (i-1)\delta) + cos(90-i\delta)= \\
& cos(\frac{360+(1-2i)\delta}{2})cos(\frac{180+\delta}{2}) =  \\ 
& cos(180 - \frac{2i-1}{2}\delta)cos(90+\frac{\delta}{2})= \\ 
& cos(\frac{2i-1}{2}\delta)sin(\frac{\delta}{2}) \\
\text{y}: \quad & \frac{1}{2}(sin(270-(i-1)\delta)+sin(90-i\delta)= \\
& sin(\frac{360-(2i-1)\delta}{2})cos(\frac{180+\delta}{2})= \\ 
& sin(180-\frac{2i-1}{2}\delta)sin(-\frac{\delta}{2})= \\
& -sin(\frac{2i-1}{2}\delta)sin(\frac{\delta}{2}) 
\end{align*}
Normalizing gives:
\begin{align*}
    \sqrt{(cos(\frac{2i-1}{2}\delta)sin(\frac{\delta}{2}))^2 + (-sin(\frac{2i-1}{2}\delta)sin(\frac{\delta}{2}))^2} = 
    sin(\frac{\delta}{2})\sqrt{cos(\frac{2i-1}{2}\delta)^2 + sin(\frac{2i-1}{2}\delta)^2} = 
    sin(\frac{\delta}{2})
\end{align*}
The normalized vector is:
\begin{align*}
    (cos(\frac{2i-1}{2}\delta), -sin(\frac{2i-1}{2}\delta)) = 
    (cos(-\frac{2i-1}{2}\delta), sin(-\frac{2i-1}{2}\delta))
\end{align*}
Since $\alpha=0$, according to Eq. \eqref{eq:Delta_closed_form},this gives $\Delta(x_i)$. 

We now calculate the diversity for each user $i < N-1$.
User $i=1$ has items $x_1$ and $x_2$ in his list. Thus, diversity is:
\begin{align*}
    \frac{1-x_1^Tx_2}{2} = 
    \frac{1}{2}(1-(1 \cdot (-cos(\frac{4-1}{2}\delta) + 0 \cdot sin(\frac{4-1}{2}\delta))) = 
    \frac{1 + cos(\frac{3\delta}{2})}{2}
\end{align*}
For even $i$, we first calculate the cosine similarity:
\begin{align*}
    & x_i^Tx_{i+1} = \\
    & -cos(\frac{2i-1}{2}\delta)cos(-\frac{2(i+1)-1}{2}\delta) + sin(\frac{2i-1}{2}\delta)sin(-\frac{2(i+1)-1}{2}\delta) = \\
    & -cos(\frac{2i-1}{2}\delta)cos(-\frac{2i+1}{2}\delta)+  sin(\frac{2i-1}{2}\delta)sin(-\frac{2i+1}{2}\delta) = \\
    & -[cos(\frac{2i-1}{2}\delta)cos(-\frac{2i+1}{2}\delta)-  sin(\frac{2i-1}{2}\delta)sin(-\frac{2i+1}{2}\delta)] = \\
    & -cos(\frac{2i-1}{2}\delta -\frac{2i+1}{2}\delta) = \\
    & -cos(\delta)
\end{align*}
Diversity is given by:
\begin{align*}
& \frac{1-x_i^Tx_{i-1}}{2} = \frac{1+cos(\delta)}{2}    
\end{align*}
For odd $i$, cosine similarity is:
\begin{align*}
    & x_i^Tx_{i+1} = \\
    & -cos(-\frac{2i-1}{2}\delta)cos(\frac{2(i+1)-1}{2}\delta) + sin(-\frac{2i-1}{2}\delta)sin(\frac{2(i+1)-1}{2}\delta) = \\
    & -cos(-\frac{2i-1}{2}\delta)cos(\frac{2i+1}{2}\delta)+  sin(-\frac{2i-1}{2}\delta)sin(\frac{2i+1}{2}\delta) = \\
    & -[cos(-\frac{2i-1}{2}\delta)cos(\frac{2i+1}{2}\delta)-  sin(-\frac{2i-1}{2}\delta)sin(\frac{2i+1}{2}\delta)] = \\
    & -cos(-\frac{2i-1}{2}\delta +\frac{2i+1}{2}\delta) = \\
    & -cos(\delta)
\end{align*}
Diversity is given by:
\begin{align*}
& \frac{1-x_i^Tx_{i-1}}{2} = \frac{1+cos(\delta)}{2}    
\end{align*}
We have remaining diversity for users $N-1$ and $N$.
Denote their corresponding diversities by $d_{N-1}$ and $d_N$.
The overall average diversity (Eq. \eqref{eq:divk}) is:
\begin{align*}
    \diversity = \frac{1}{N}\left(d _{N-1} + d_N +\frac{1 + cos(\frac{3\delta}{2})}{2} + \frac{(N-3)(1+cos(\delta))}{2}\right) 
\end{align*}
Since $d_{N-1}$ and $d_N$ are non-negative, we get:
\begin{align*}
\diversity \ge \frac{1}{N} \left(\frac{1 + cos(\frac{3\delta}{2})}{2} + \frac{(N-3)(1+cos(\delta))}{2}\right)
\end{align*}
Note $\frac{1 + cos(\frac{3\delta}{2})}{2}$ is also non-negative since $cos(\frac{3\delta}{2})$ cannot be less than $-1$.
Hence, we get:
\begin{align*}
\diversity \ge \frac{(N-3)(1+cos(\delta))}{2N} = \frac{N-3}{N}\frac{1+cos(\delta)}{2}
\end{align*}
Finally, set $\delta = cos^{-1}(1-2\epsilon)$; this gives $\epsilon = \frac{1-cos(\delta)}{2}$, and since $1-\epsilon = \frac{1+cos(\delta)}{2}$, we get as required:
\begin{align*}
\diversity \ge \left(1-\frac{3}{N}\right)(1-\epsilon)
\end{align*}
\endproof

\section{Additional Experimental Results: Synthetic Data}
\label{apx:additional_result}


\begin{figure}[t!]
    \centering
    \includegraphics[width=0.45\linewidth]{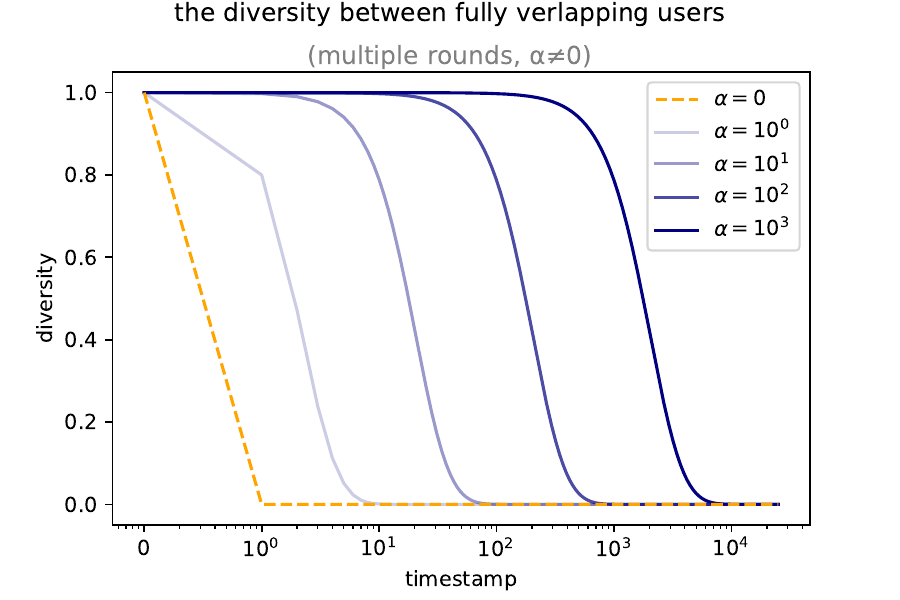}
    \caption{
    Empirical generalization of Proposition \ref{prop:Full_users_overlap} for $\alpha \ne 0$.
    }
    \label{fig:overlap_emp}
\end{figure}

\subsection{Empirical analysis: Proposition \ref{prop:Full_users_overlap} beyond $\alpha = 0$}
\label{apx:empirical_generalization_full_overlapping}
Proposition \ref{prop:Full_users_overlap} states that full user 
overlap leads to zero diversity, but relies on the assumption that $\alpha=0$, i.e., that there are no modification costs, and so items can move arbitrarily on the unit sphere.
Our motivation for considering $\alpha=0$ (and a single time step)
was that it is a useful proxy for larger $\alpha$ over multiple updates (that do not include retraining).
Here we demonstrate empirically that this is indeed the case,
i.e., that for $\alpha>0$, diversity does go to zero over time.

We consider two items, positioned so that they are as far apart as possible: $x_1=(1,0)$ and the other at $x_2=(-1, 0)$.
We set $\vtilde$, which represents the spherical average of the overlapping users, to a random  direction. We then measure how diversity changes over the time. As shown in Figure
\ref{fig:overlap_emp}
diversity drops to zero for all $\alpha$ considered.



\subsection{Accuracy and diversity over time}
\label{sec:multi_rounds}
\begin{figure*}[t!]
\label{fig:synth_round}
\centering
\includegraphics[width=0.4\textwidth]{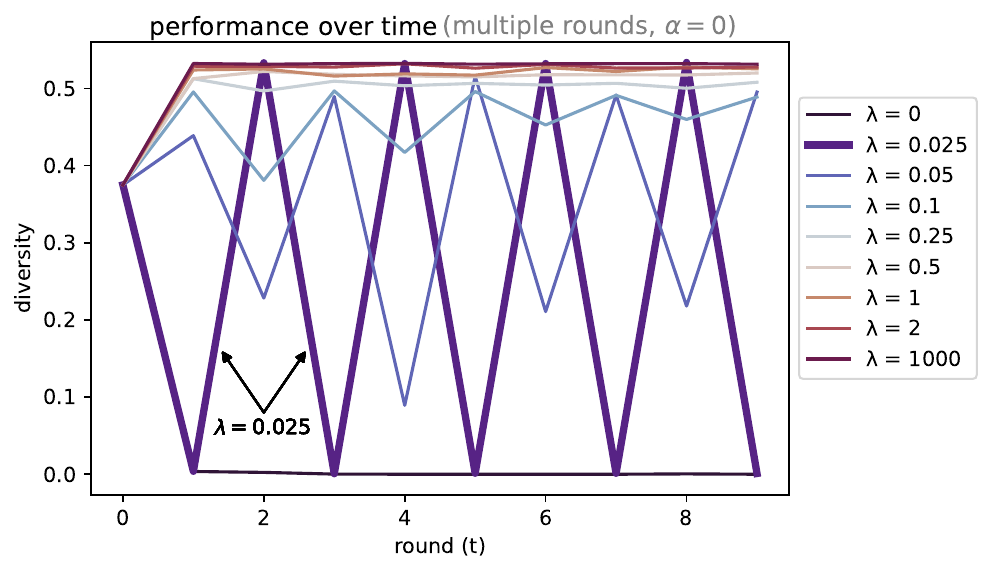} \qquad
\includegraphics[width=0.4\textwidth]{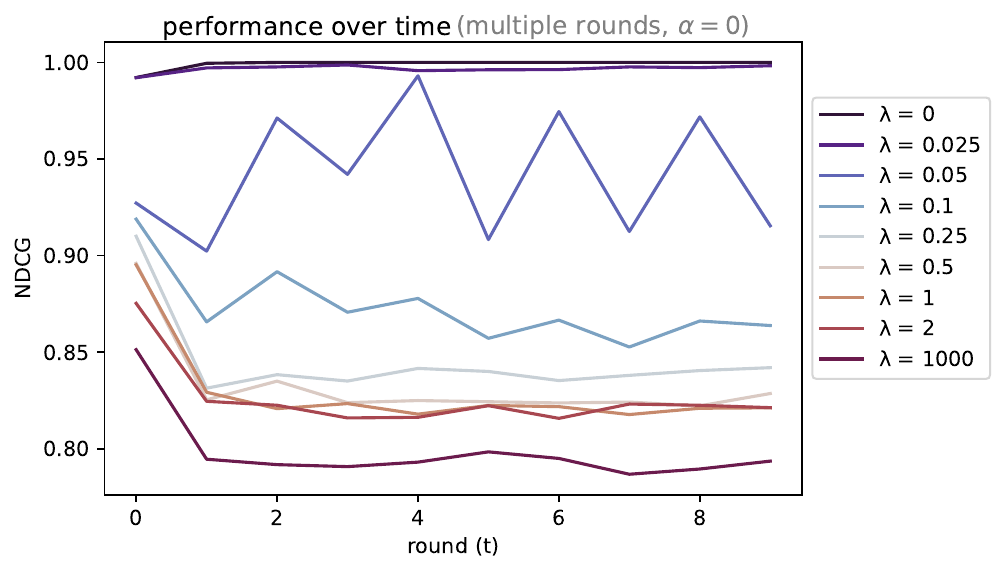}
\caption{\textbf{(Left)} diversity vs number of rounds. \textbf{(Right)} NDCG vs number of rounds.}
\label{fig:synth time}
\end{figure*}

As an intermediate step between the experiments in Sec. \ref{variation_u*} and Sec. \ref{sec:synth_vary_costs},
here we consider how accuracy and diversity trade off over time and for a range of $\lambda$, but while keeping $\alpha=0$.
Results are shown in Figure \ref{fig:synth_round}. For $\lambda=0$ (i.e., learning does not regularize for diversity), resulting NDCG is high and fixed, but diversity is nearly zero. This can be explained by the small value of $\sigma_{u^*}$, which implies that after strategic modification, items will be highly similar---which eliminates diversity (as in experiment \ref{variation_u*}), and makes the ranking task easy (since all items are similarly relevant).
In the other extreme, when $\lambda = 1000$ (and so diversity is heavily regularized for), we observe that in the first round, diversity increases sharply, whereas NDCG decreases.
In the subsequent rounds, diversity remains high and NDCG remains low.

For intermediary $\lambda$, results show how the learning allows to balance NDCG and diversity, where diversity is obtained for little loss in NDCG. For larger $\lambda$, more NDCG is sacrificed and diversity increases. An interesting phenomena is that lower values of $\lambda$ exhibit periodic behavior.
For example, for $\lambda=0.025$, diversity alternates between very high (even round) and very low (odd rounds).
Given how NDCG and diversity relate, we see two possible explanations for this:
\begin{enumerate}[noitemsep,leftmargin=1cm,label=(\arabic*)]
\item
When diversity is high,
the relevance of items may differ significantly. As a result, 
even a mild change in the item's ranking can cause NDCG to drop. To avoid this loss of NDCG, one solution is for the model to learn user embeddings $u_i$ that are close to $u^*$. This, however, causes diversity to be low in the next round.
\item When diversity is low, item features are very similar, and hence have highly similar relevance values. This implies that most rankings will have similar NDCG.
Consequently, learning can encourage diversity without sacrificing current NDCG.
\end{enumerate}

\subsection{The role of the number of recommended items $k$}
\begin{figure*}[t!]
\centering
\includegraphics[width=0.4\textwidth]{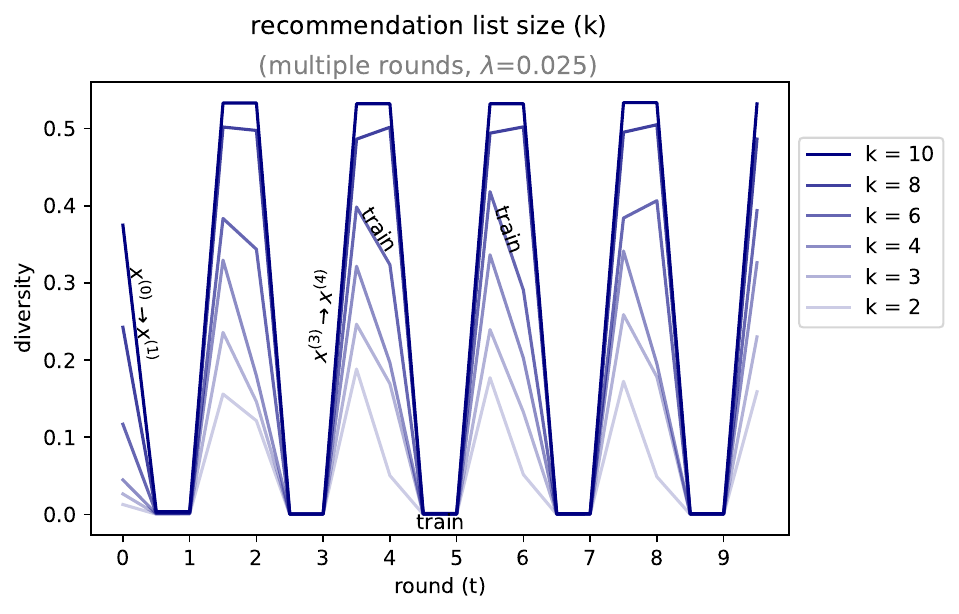} \qquad
\includegraphics[width=0.4\textwidth]{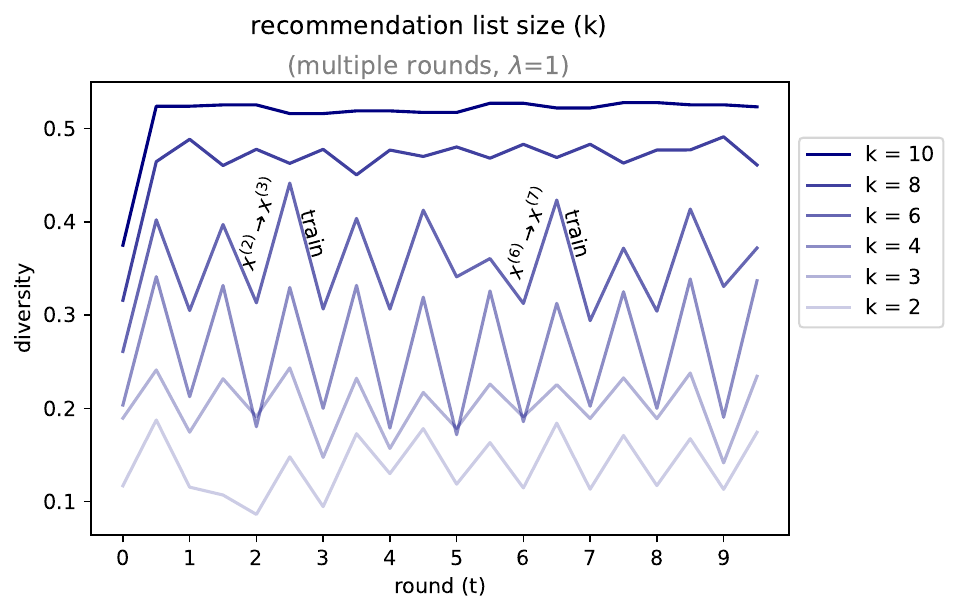}
\caption{\textbf{(Left)} diversity vs number of rounds for $\lambda=0.025$. \textbf{(Right)} diversity vs number of rounds for $\lambda=1$.}
\label{fig:synth k time}
\end{figure*}

Here we consider how the size of the recommendation list $k$ impacts the ability of the model to create diversity. We keep all other aspects of the experiment fixed, and vary $k$. Figure \ref{fig:synth k time} shows results for two values of $\lambda$: a low (but positive) value of $\lambda=0.025$ (left), and a high value of $\lambda=1$.
Overall, results show that higher $k$ enables larger diversity; conversely, when $k$ is small, results show that it is harder for the model to encourage diversity.
One possible reason is that due to linearity, the top of the list is likely to includes items that are similar.
The main distinction between the low and high $\lambda$ is how diversity appears over time. As in the previous experiment in Sec. \ref{sec:multi_rounds} (in which $k=10$),
we see that $\lambda=0.025$ exhibits alternating diversity for all $k$.
For $\lambda=1$, lower $k$ still exhibits some fluctuations, but these become small as $k$ grows.



\section{Experimental details}
\label{apx:Data details}
\subsection{Data} 

Our experiments use the Yelp dataset,
which is publicly available at \url{https://www.yelp.com/dataset/download}.\footnote{Note Yelp periodically updates their repository; to ensure consistency, we include in our code preprocessed data, used in our experiments, that was parsed from raw data published by Yelp on July 2021.}

\paragraph{Items.}
Yelp includes data about many business types;
of these, our experiment focuses on restaurants.
To obtain restaurant entries,
we manually identify and select all categories that pertain to restaurants (e.g, `pizzaria' or `burger bar').
This results in 22,197 distinct entries.

\paragraph{Features.}
For features, we use a subset of the available features that were prevalent, and which were found to be informative for training $f^*$.
We use category information to form additional features by grouping similar categories having similar contextual meaning;
for example, the categories `pizza', `pasta', `calzone', etc. are assigned the binary feature `Italian cuisine'.
Overall we use 43 features, which include: 

'stars', 'alcohol', 'restaurants good for groups', 'restaurants reservations', 'restaurants attire', 'bike parking', 'restaurants price range', 'has tv', 'noise level', 'restaurants take out', 'caters', 'outdoor seating', 'good for meal-dessert', 'good for meal-late night', 'good for meal-lunch', 'good for meal-dinner', 'good for meal-brunch', 'good for meal-breakfast', 'dogs allowed', 'restaurants delivery', 'japanese', 'chinese', 'india', 'middle east', 'mexican food', 'sweets', 'coffee', 'italian', 'burgers', 'hot dogs', 'sandwiches', 'steak', 'pizza', 'seafood', 'fast food', 'vegan', 'ice cream', 'restaurants table service', 'business accepts credit cards', 'wheel chair accessible', 'drive thru', 'happy hour', 'corkage'.

\paragraph{Users.}
As noted, we focus on active users who have contributed at least 100 reviews.
One reason is that we have found that including low-activity users in this dataset results in a sparse and disconnected graph,
with many isolated items (and corresponding users);
this trivializes the task of diversification since most items can be incentivized independently.
For each user, we consider the 40 most popular items (of those rated by that user),
since by similar reasoning these provide larger overlap, and hence more intricate dependencies across items.
In particular, we use the following procedure to construct potential item lists:

Denote by $R$ the set of all restaurants, and by $U$ set of active users.
\begin{itemize}[label={\tiny{$\bullet$}}]
    \item for all users $u \in U$, initialize $X_u = \emptyset$
    \item for all restaurants $r \in R$, initialize $U_r$ to include all users $u$ that have reviewed restaurant $r$ 
    \item while exists $u \in U$ for which $|X_u| < 40$:
    \begin{itemize}[label={\tiny{$\bullet$}}]
        \item let $r$ be the restaurant with the most users in $U$, i.e., $U_r \cap U$ is largest
        \item for each user $u \in U_r$: 
        \begin{itemize}[label={\tiny{$\bullet$}}]
            \item add $r$ to $X_u$
            \item remove $r$ from $R$
            \item if $|X_u|=40$, then for each $r'$ in $x_u$, remove $u$ from $U_{r'}$
        \end{itemize}
    \end{itemize}
\end{itemize}

\subsection{Generating counterfactual ground-truth labels (pre-processing)}
\label{apx:training f^*}

Since our experiments include modified items $x^f_j$ that do not exist in the data, for learning and evaluation we require means to generate corresponding counterfactual relevance scores $y^f_j$.
To achieve this, prior to the experiment we train a ground truth labeling function, $f^*(x)$, which we query for updated labels throughout the experiment. As noted, we ensure $f^*$ is distinct from (and more powerful than) the predictive models $f$ we learn in the actual experiment.

\paragraph{Data for training $f^*$.}
We train $f^*$ on all data generated by users having at least 50 restaurant reviews; these amount to 1,377 users and 113,852 reviews. 
Since the original data does not include informative user features,
we generate ground-truth user features $u^*_i$ by aggregating for each user $i$ the features of all restaurants reviewed by $i$.
This is similar in spirit to approaches for content-based recommendation.
Formally, we define
$u_i^* = \frac{1}{{|\rev_i|}}\sum_{x\in \rev_i} x$,
where $\rev_i$ is the set of all restaurants that the user $i$ reviewed.

For labels, we consider probabilistic labels $y_{ij} \in [0,1]$ that describe the likelihood that user $i$ will visit restaurant $j$,
which we interpret as relevance.
For each user $i$, we set $y_{ij}=1$ for every restaurant $j$ that $i$ reviewed.
To obtain negative labels, for each $(i,j)$ pair, we first obtain the geographical location of restaurants $j$, and then retrieve the closest restaurant $j'$ (in geographical terms) to $j$ that $i$ did \emph{not} review; we then set $y_{i j'} = 0$.
This is intended to mimic a setting in which $i$ \emph{could} have went to either $j$ or $j'$ (since they are physically nearby), but chose to go to $j$.

\paragraph{Architecture.}
We set $f^*$ to be an MLP with ReLU activations.
We use five layers, which we have found to be sufficient for expressing the non-linear relations between user and item features found in the data.
The first layer has 86 inputs (43 restaurant features and 43 user features) and $86 \times 2 = 172$  outputs,
and for each consecutive layer the output dimension reduces by half.

\paragraph{Training and evaluation.}
We split the data into train, validation, and test sets,
using a 70-20-10 split.
We optimize using Adam with a learning rate of 0.01,
which gives a reasonable balance between performance and runtime,
and used the validation set for early stopping.
The final $f^*$ achieves 72\% accuracy on the held-out test set.


\subsection{Hyper-parameters and tuning (main experiment)}
For optimizing predictive models $f$ in each experimental condition,
we use Adam and train for a maximum of 200 epochs with learning rate 0.1.
For smoothing (see Sec. \ref{sec:optimization}),
we use temperatures $\tau=0.1$ for NDCG, $\tau=1$ for the permutation matrix approximation, and $\tau=5$ for the soft-$k$ function;
all were chosen to be the largest feasible values that permit smooth training.
All experiments were run on a cluster of AMD EPYC 7713 machines (1.6 Ghz, 256M, 128 cores). 

 
\section{Additional experimental results: real data}
\label{apx:results_real}

\subsection{Diversity over time -- additional results} \label{apx:additional_div}
\begin{figure*}[h!]
\centering
\includegraphics[width=0.24\textwidth]{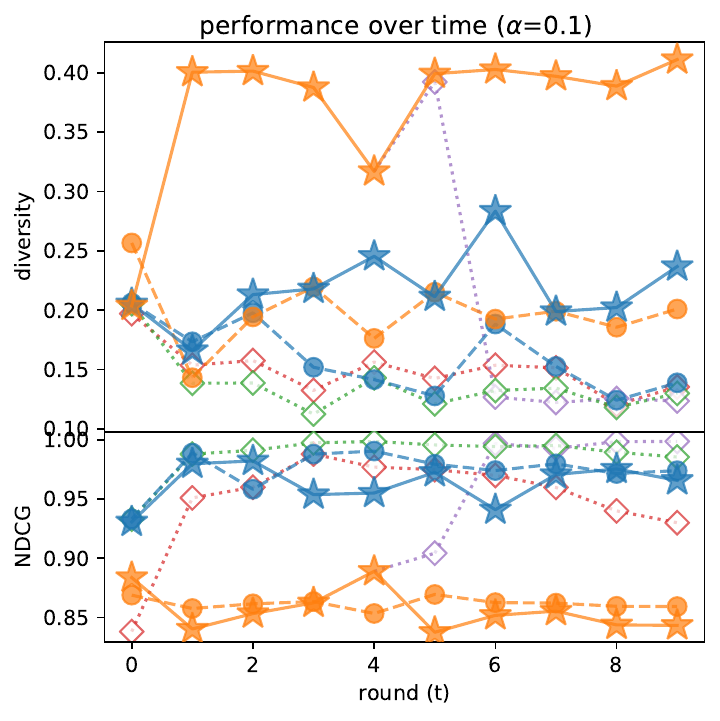}
\includegraphics[width=0.24\textwidth]{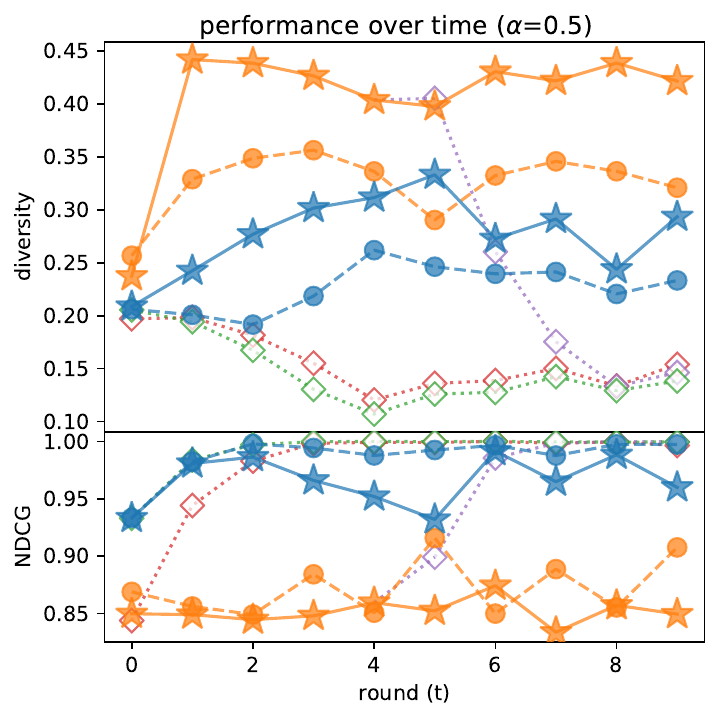}
\includegraphics[width=0.24\textwidth]{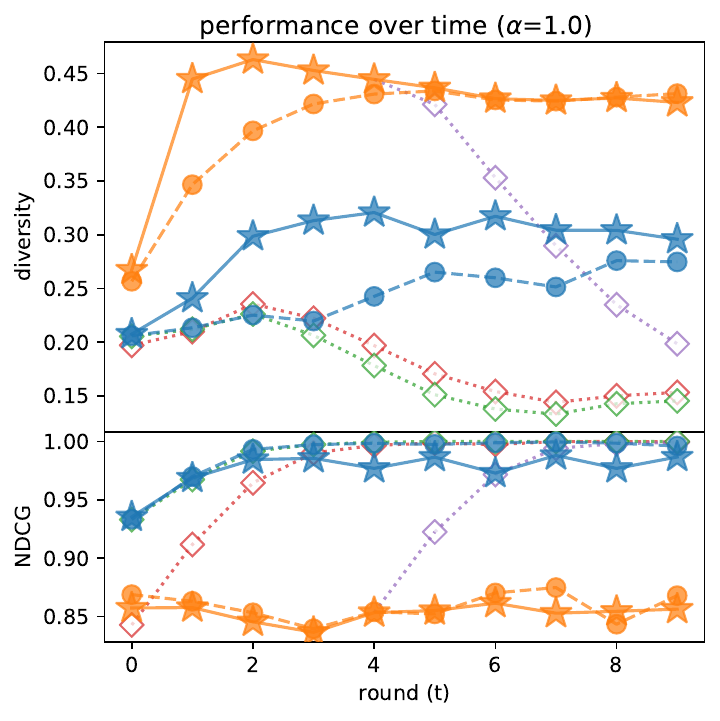}
\includegraphics[width=0.24\textwidth]{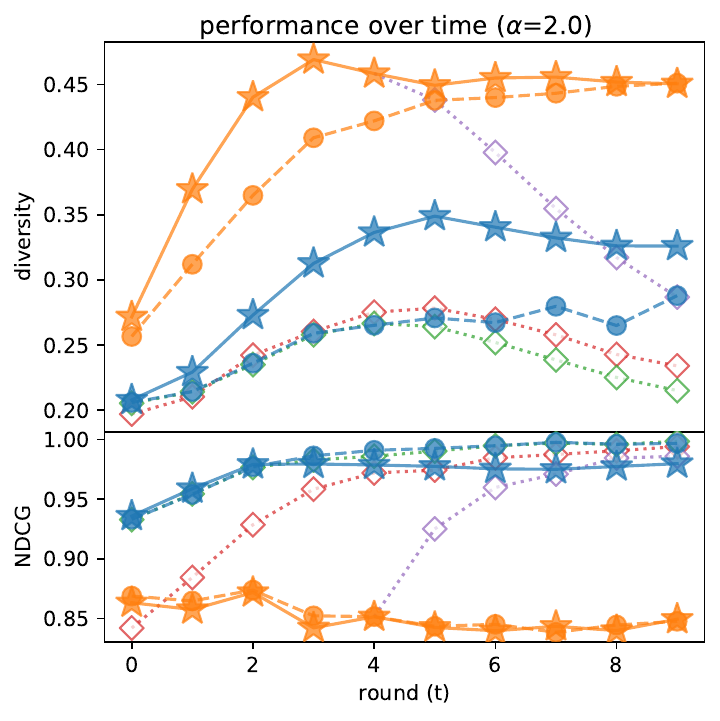}
\caption{Diversity and NDCG over rounds for different methods and target NDCG values, and for costs $\alpha \in \{0.1, 0.5, 1, 2\}$.}
\label{apx:fig:div_over_time}
\end{figure*}

Figure~\ref{apx:fig:div_over_time} includes extended results pertaining to our main experiment
in Sec. \ref{sec:div_over_time} for additional cost scales $\alpha$.


\subsection{Tradeoffs over time -- additional results} \label{apx:additional_tradeoff}
\begin{figure}[h!]
    \centering
    \includegraphics[width=0.24\columnwidth]{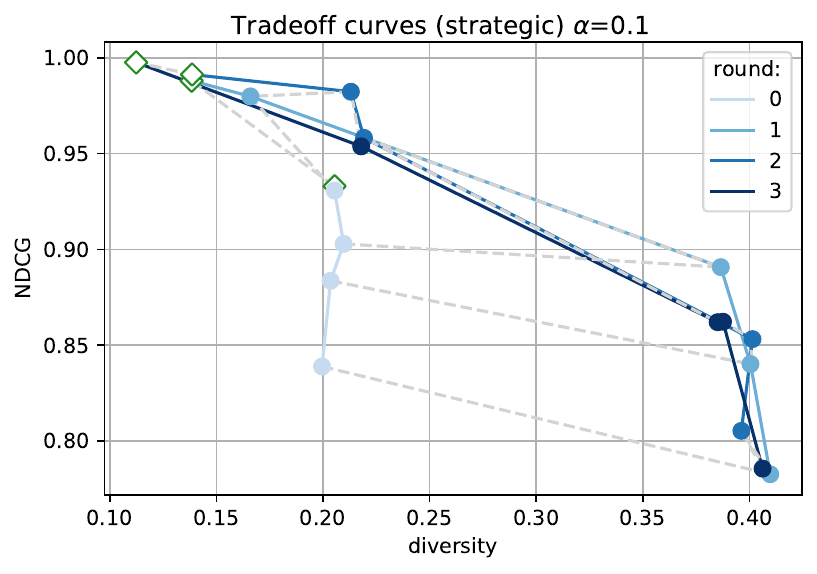}
    \includegraphics[width=0.24\columnwidth]{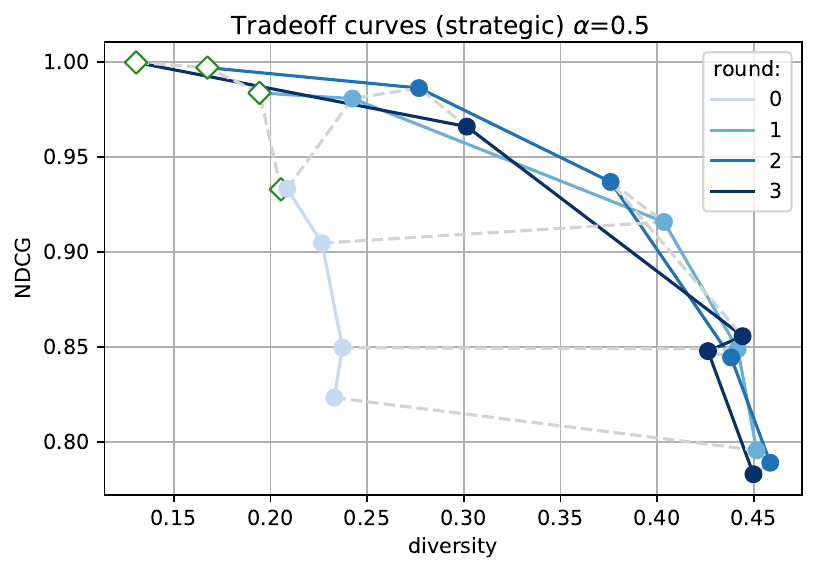}
    \includegraphics[width=0.24\columnwidth]{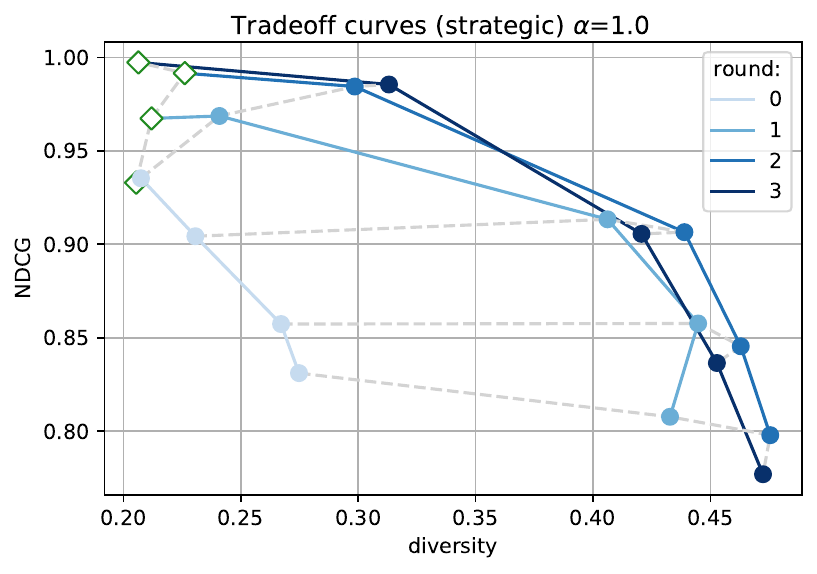}
    \includegraphics[width=0.24\columnwidth]{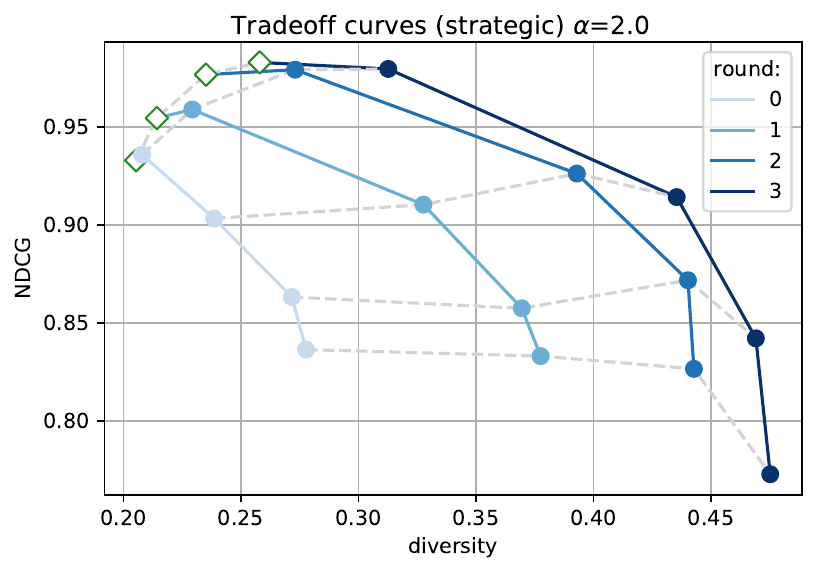}
    \caption{
    Pareto curves for \method{strategic} per round (varying $\lambda$; solid lines), and vs. \method{baseline}.
    Dashed lines show per-$\lambda$ trajectories.\looseness=-1
    }
    \label{apx:fig:tradeoff}
\end{figure}

Figure~\ref{apx:fig:tradeoff} includes extended results for our experiment on tradeoffs over time in Sec. \ref{sec:tradeoff} for additional cost scales $\alpha$.

\subsection{Sensitivity to a misspecification of the response model}
\label{apx:sensitivity}
Our experiments in Sec.~\ref{sec:real data} consider a setting in which the system has knowledge of the response model $\Delta$,
and in particular, of the true cost scale $\alpha$.
In this section we explore the sensitivity of our approach to learning under misspecified $\alpha$.
In particular, in each experimental instance, we train our model on some $\alpha$, but test it on a different $\alpha'$.
Note that the misspecified $\alpha$ is used throughout all training rounds, and so the effects of misspecification accumulate.



\begin{figure}[h!]
\centering
\includegraphics[width=0.35\textwidth]{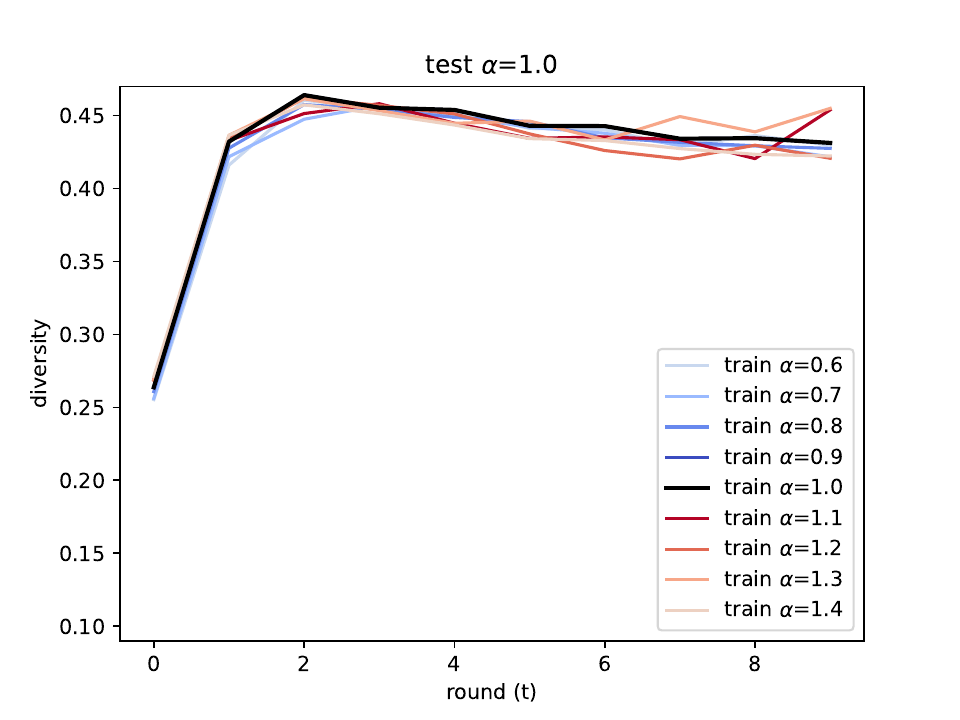} \quad
\includegraphics[width=0.35\textwidth]{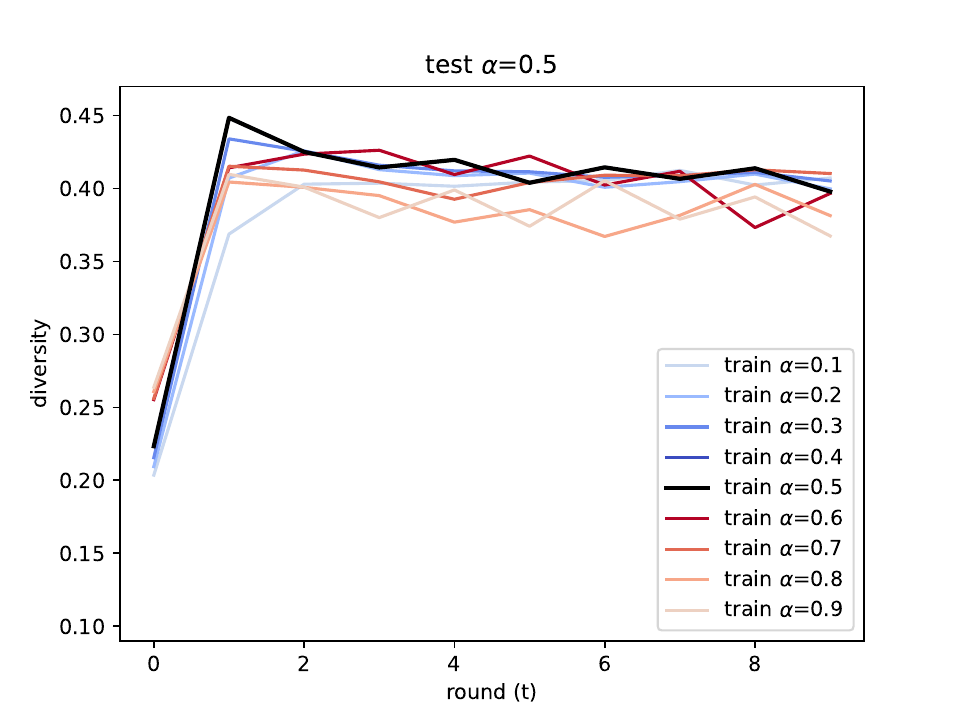}
\caption{Diversity for misspecified cost scale $\alpha$, for fixed test $\alpha$ and varying train $\alpha$.}
\label{apx:fig:sensitivity_test}
\end{figure}

Figure~\ref{apx:fig:sensitivity_test} (left) shows diversity over rounds 
on a fixed test $\alpha=1$, for smaller training $\alpha \in [0.6,1)$ (blue lines), larger training $\alpha \in (1,1.4]$ (red lines), and the correct training $\alpha=1$ (black line).
Results show our approach is fairly robust to misspecification,
with performance for all train $\alpha$ almost matching the correct one.
Figure~\ref{apx:fig:sensitivity_test} (right) shows similar results for test $\alpha=0.5$. 
Here robustness is preserved in full for smaller $\alpha \in [0.1,0.5)$,
but shows some deterioration in performance for the larger $\alpha \in (1,1.4]$.

\begin{figure}[h!]
\centering
\includegraphics[width=0.35\textwidth]{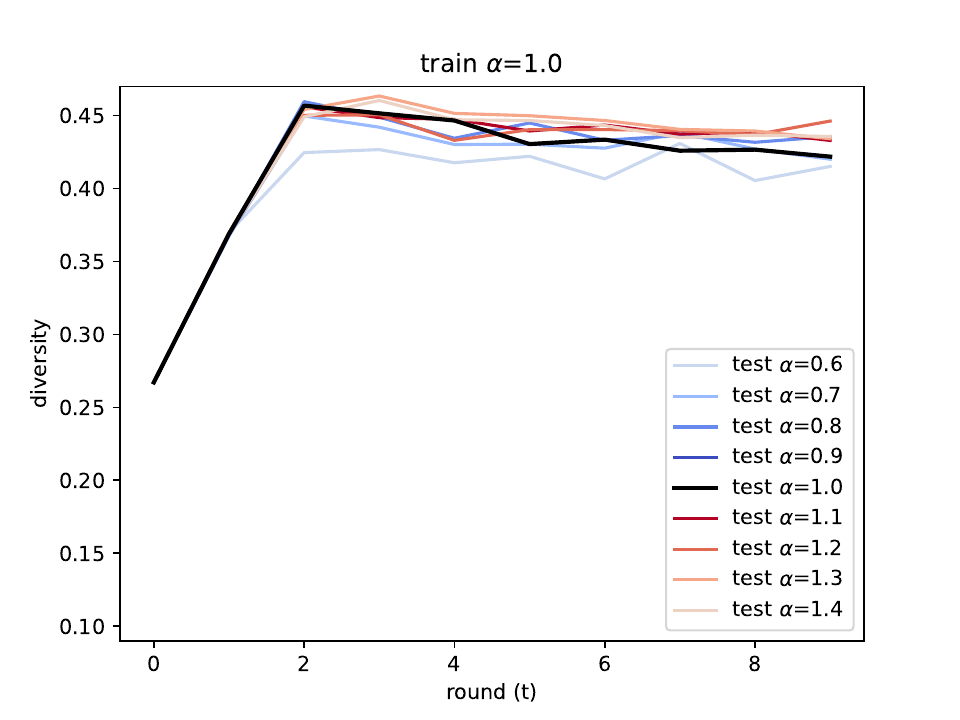} \quad
\includegraphics[width=0.35\textwidth]{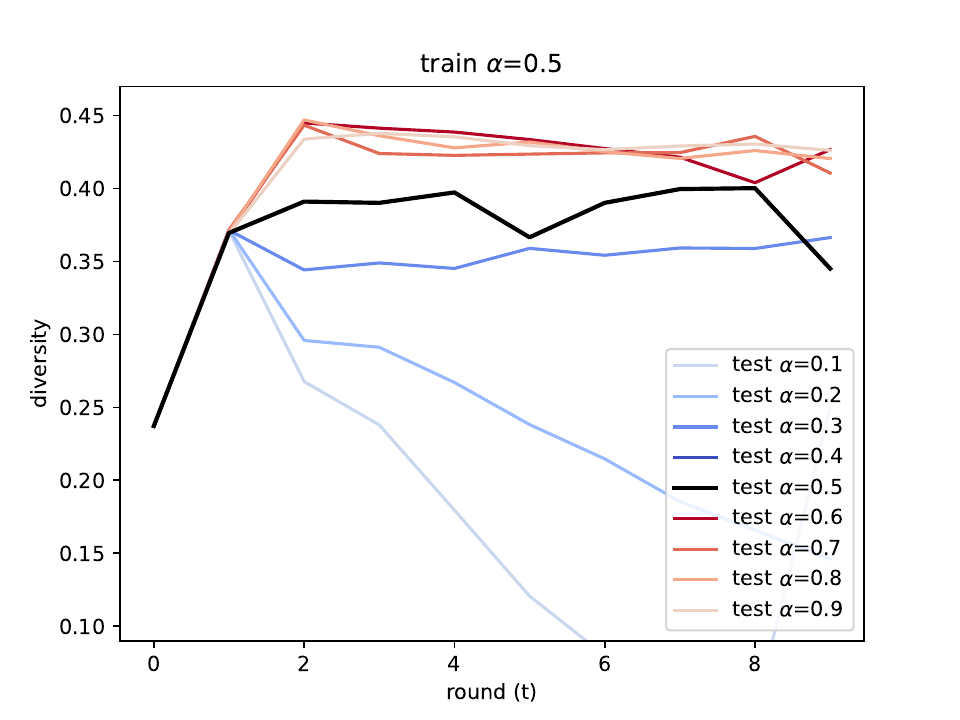}
\caption{Diversity for misspecified cost scale $\alpha$, for fixed train $\alpha$ and varying test $\alpha$.}
\label{apx:fig:sensitivity_train}
\end{figure}

Complementarily, Figure~\ref{apx:fig:sensitivity_train} shows diversity for fixed train $\alpha$ and varying test $\alpha$.
Here, performance is again robust for $\alpha=1$ (left).
However, for the smaller train $\alpha=0.5$ (right), 
in which items are subject to more dramatic modifications,
mispecification has a significant effect on performance.
For smaller test $\alpha$ (blue lines), sever overestimation of $\alpha$ in training (e.g., 0.5 vs. test $\alpha=0.1$)
has a severe negative effect on diversity over time.
Interestingly, underestimation of $\alpha$ in training (red lines) results in improved diversity, suggesting that perhaps taking excessive cautionary steps is helpful in this case.

\subsection{Entropy-based diversity regularization} \label{apx:entropy}
As we state in Sec.~\ref{sec:Setup}, our paper focuses predominantly on cosine similarity, which we believe is appropriate for the recommendation environment we consider, and is a popular choice in the literature.
Nonetheless, our approach is not restricted to this choice, and in this section we describe how it can be extended to operate on other similarity measures, and in particular, on entroty-based similarity.
We then provide some empirical results for this settings.


To begin, note that
conventional entropy regularization (e.g., \citet{qin2013promoting}) assumes a Gaussian distribution over feature vectors, and so is not immediately applicable to our setting of unit-norm features. To account for this, we propose a similar measure, but based on the Beta distribution, which is appropriate for inputs in [0,1], and which can apply per feature.
The benefits of this measure are that:
(i) its parameters can be efficiently estimated using moment matching \citep{owen2008parameter};
(ii) both parameter estimates and the differential entropy function are differentiable, and hence permit gradients to pass through; and
(iii) entropy can be made to take strategically-modified inputs, and hence allow for strategically-aware optimization.

The Beta distribution is defined by two shape parameters, $a>0$ and $b>0$.
Let $Z\sim\textrm{Beta}(a,b)$,
and for a given sample of such $Z$-s,
denote its average by $\Bar{Z}$ its standard deviation by $S$.
Then the parameters $a$ and $b$ can be efficiently estimated as:
\[
\hat{a} = \Bar{Z}(\Bar{Z}\frac{(1-\Bar{Z})}{S^2}-1), \qquad \qquad
\hat{b} = (1-\Bar{Z})(\Bar{Z}\frac{(1-\Bar{Z})}{S^2}-1)
\]
Our approach is to consider each feature in each item list as deriving from some Beta distribution.
Hence, for a given list $X$ and feature $i$, we first estimate
$\hat{a},\hat{b}$ using $\{x_i\}_{x \in X}$.
Note that both estimands are differentiable.
Then, we compute entropy for this list and feature as
$e_i(X) = \textrm{entropy}(\hat{a},\hat{b})$, which admits a differentiable closed form (we used the pytorch implementation\footnote{\url{https://pytorch.org/docs/stable/distributions.html}} that allows to pass gradients).
Finally, we define $\diversity(X) = \frac{1}{d}\sum_{i=1}^d e_i(X)$.
We can then replace $X$ with $X^f$, and plug into our objective in Eq.~\eqref{eq:learning_objective_strat}, which remains differentiable.

\begin{figure*}[h!]
\centering
\includegraphics[width=0.24\textwidth]{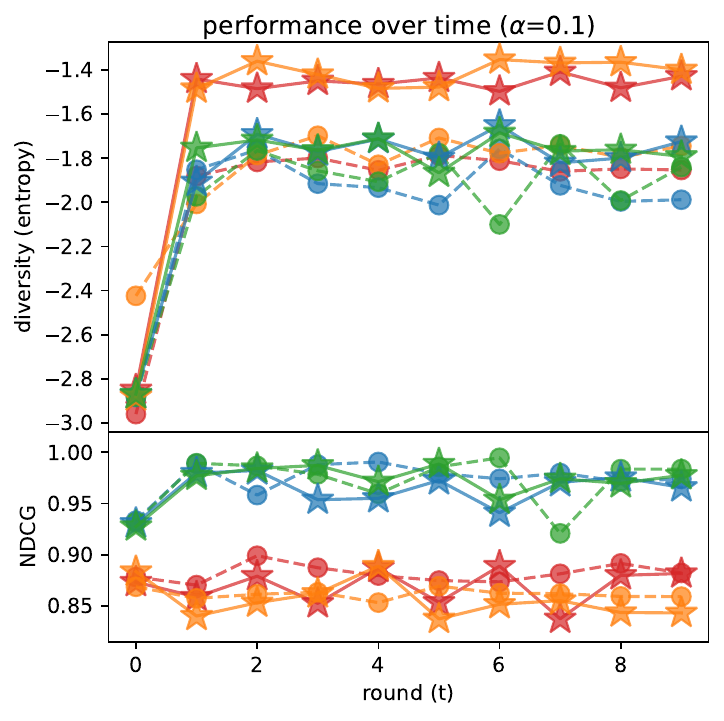}
\includegraphics[width=0.24\textwidth]{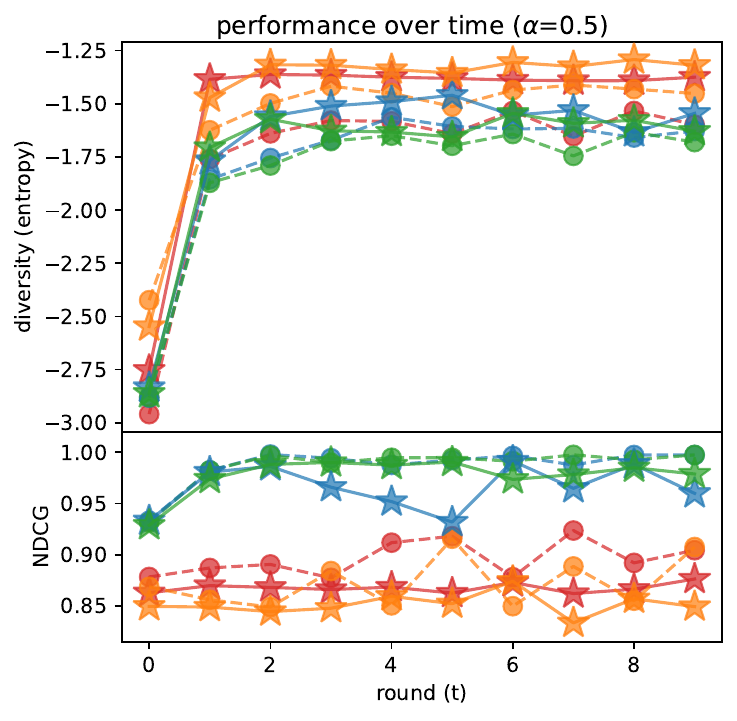}
\includegraphics[width=0.24\textwidth]{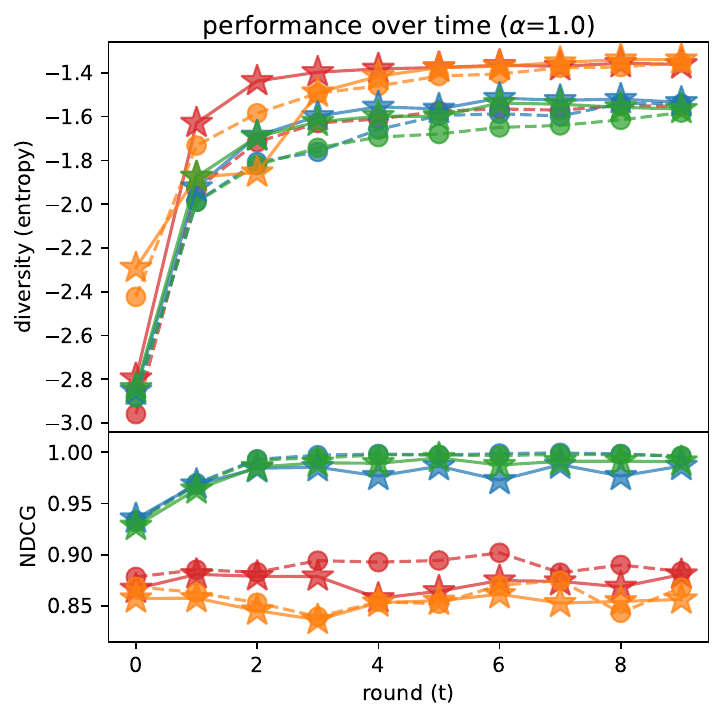}
\includegraphics[width=0.24\textwidth]{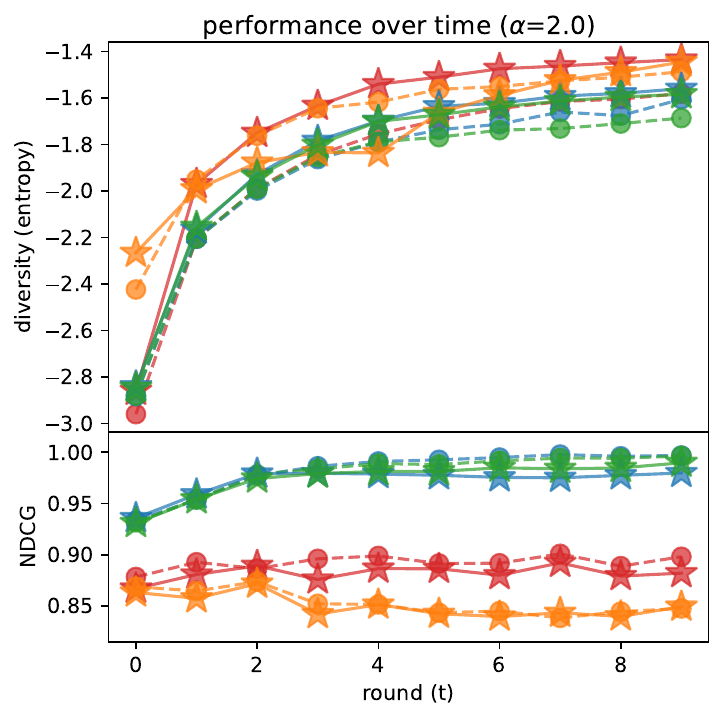} \\
\includegraphics[width=0.6\textwidth]{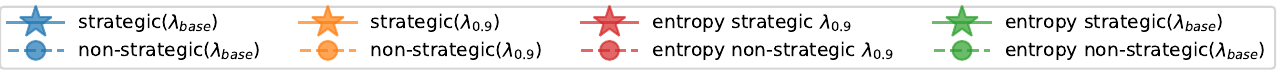}
\caption{Experiments using entropy-based similarity as a diversity measure.}
\label{apx:fig:entropy_reg_test_entropy}
\end{figure*}

Using this approach, we extend our main experiments to also include entropy-based regularization.
Figure~\ref{apx:fig:entropy_reg_test_entropy} shows diversity and NDCG for all methods, when diversity is measured using entropy.
Here again we see that strategically-aware methods outperform non-strategic methods across multiple cost scales $\alpha$, although to a lesser extent than when measuring cosine similarity.
Interestingly, optimizing for the incorrect correct measure
(here, cosine; blue and orange) performs as well as when the correct measure (i.e., entropy; red and green) is optimized.


Finally, we rerun our original experiment using cosine similarity as a measure of diversity, but considering also methods that optimize entropy-based similarity (red and green).
Here we see that misspecified diversity regularization is useful, but to a lesser extent than the correct form of regulariztaion.
Nonetheless, the importance of awareness to strategic behavior remains to be more important (in terms of performance) than applying the correct regularizer.

\begin{figure*}[h!]
\centering
\includegraphics[width=0.24\textwidth]{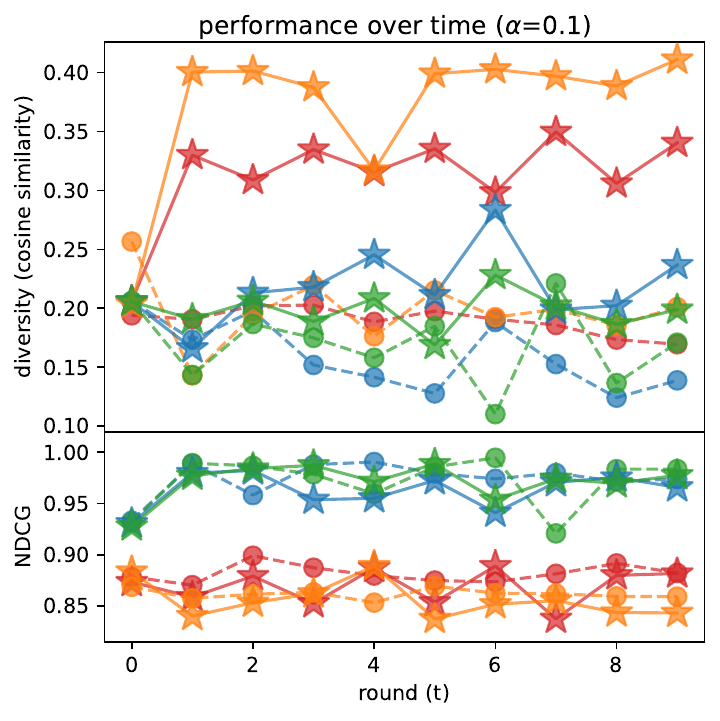}
\includegraphics[width=0.24\textwidth]{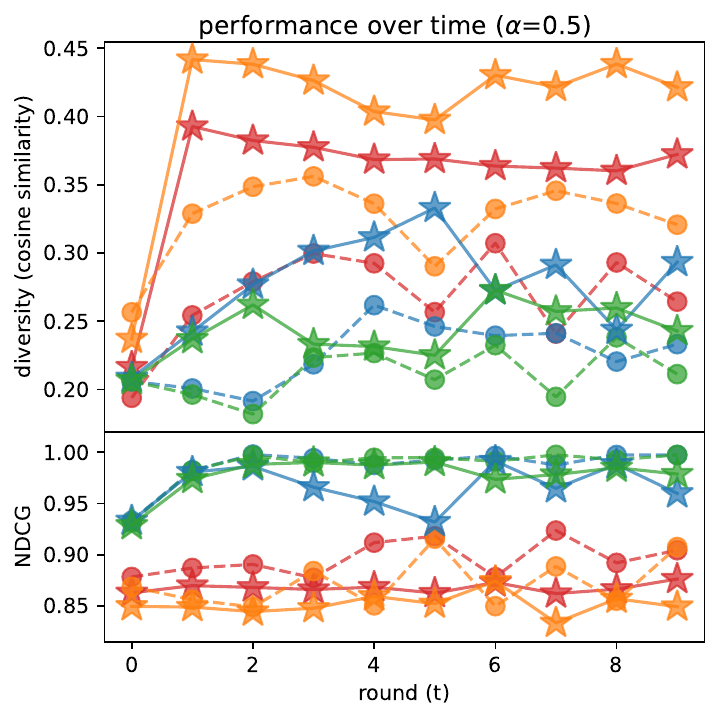}
\includegraphics[width=0.24\textwidth]{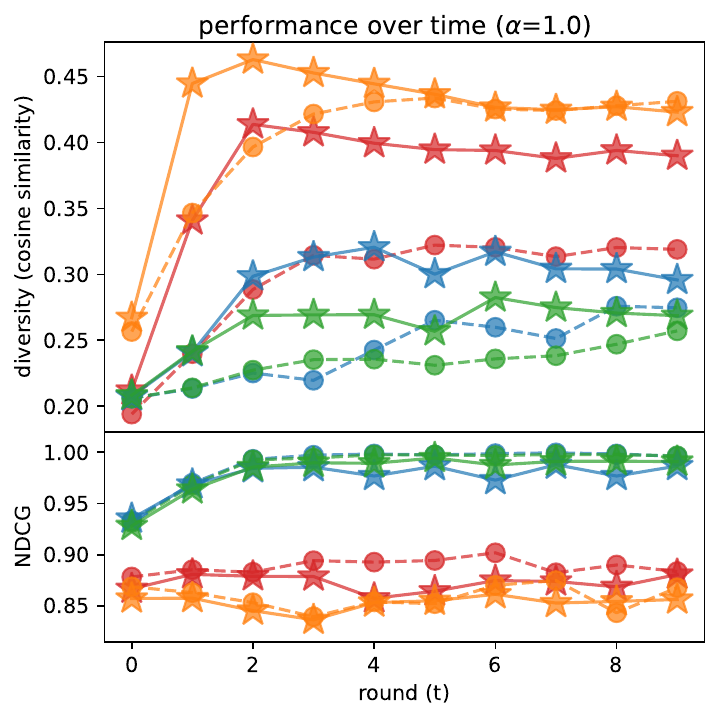} 
\includegraphics[width=0.24\textwidth]{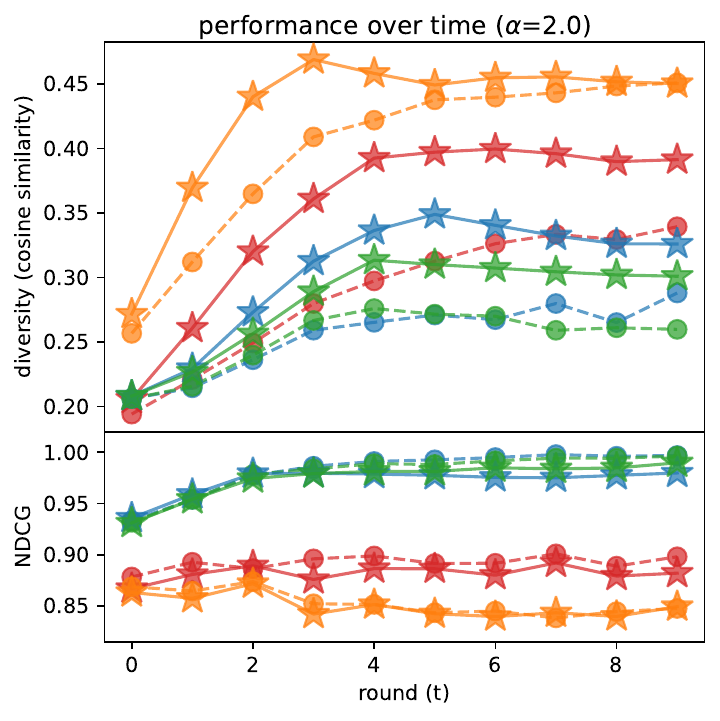} \\
\includegraphics[width=0.6\textwidth]{rebuttal_plots/cosine_vs_entropy_cosine_test/legend_over_time.pdf}
\caption{Experiments using cosine similarity as a diversity measure, but including also methods that optimize entropy-based diversity.}
\label{apx:fig:entropy_reg_test_cosine}
\end{figure*}